%% file: mixsa.tex
\documentclass[lettersize,journal]{IEEEtran}
\usepackage{amsmath,amsfonts}
\usepackage{algorithmic}
\usepackage{algorithm}
\usepackage{array}
\usepackage[caption=false,font=normalsize,labelfont=sf,textfont=sf]{subfig}
\usepackage{textcomp}
\usepackage{stfloats}
\usepackage{url}
\usepackage{verbatim}
\usepackage{graphicx}
\usepackage{multirow}
\usepackage{subcaption}
\usepackage{subfig}
\usepackage{cite}
\usepackage{booktabs}
\usepackage[colorlinks, linkcolor=blue]{hyperref}

\usepackage{xcolor}
\usepackage{xpatch}

\begin{document}

\title{MixSA: Training-free Reference-based Sketch Extraction via Mixture-of-Self-Attention}

\author{Rui Yang, Xiaojun Wu, and Shengfeng He,~\IEEEmembership{Senior Member, ~IEEE}

\thanks{This work is supported by the Guangdong Natural Science Funds for Distinguished Young Scholar (No. 2023B1515020097), the National Research Foundation Singapore under the AI Singapore Programme (AISG3-GV-2023-011), the National Natural Science Foundation of China(No. 62377034), the Fundamental Research Funds for the Central Universities of China under Grant (GK202407007), the Key Laboratory of the Ministry of Culture and Tourism (No. 2023-02), the National Natural Science Foundation of China (No.11872036), and the Innovation Team Project of Shaanxi (No. 2022TD-26). Corresponding authors: Xiaojun Wu and Shengfeng He.}
\thanks{Rui Yang is with the School of Computer Sciences, Shaanxi Normal University, Xi'an, China. E-mail: rane@snnu.edu.cn.}
\thanks{Xiaojun Wu is with the School of Computer Science and the Key Laboratory of Intelligent Computing and Service Technology for Folk Song of the Ministry of Culture and Tourism, Shaanxi Normal University, Xi'an, 710119, China. E-mail: xjwu@snnu.edu.cn.}        
\thanks{Shengfeng He is with the School of Computing and Information Systems, Singapore Management University, Singapore. E-mail: shengfenghe@smu.edu.sg.}
}

\markboth{IEEE Transactions on Visualization and Computer Graphics}%
{Yang \MakeLowercase{\textit{et al.}}: MixSA: Training-free Reference-based Sketch Extraction via Mixture-of-Self-Attention}

\maketitle

\begin{abstract}
Current sketch extraction methods either require extensive training or fail to capture a wide range of artistic styles, limiting their practical applicability and versatility. We introduce \textbf{Mix}ture-of-\textbf{S}elf-\textbf{A}ttention (MixSA), a training-free sketch extraction method that leverages strong diffusion priors for enhanced sketch perception. At its core, MixSA employs a mixture-of-self-attention technique, which manipulates self-attention layers by substituting the keys and values with those from reference sketches. This allows for the seamless integration of brushstroke elements into initial outline images, offering precise control over texture density and enabling interpolation between styles to create novel, unseen styles. By aligning brushstroke styles with the texture and contours of colored images, particularly in late decoder layers handling local textures, MixSA addresses the common issue of color averaging by adjusting initial outlines. Evaluated with various perceptual metrics, MixSA demonstrates superior performance in sketch quality, flexibility, and applicability. This approach not only overcomes the limitations of existing methods but also empowers users to generate diverse, high-fidelity sketches that more accurately reflect a wide range of artistic expressions.
\end{abstract}


\section{Introduction}
\IEEEPARstart{S}{ketch} drawing, as a unique form of simplified expression, is both a powerful tool for conveying ideas and a distinctive artistic style. Strokes are fundamental in sketch art, varying in width, direction, and pressure, which significantly contribute to the depiction of form, depth, and texture. Artists use different lines to create sketches in various styles, manifested in the thickness, angle, continuity, depth, and shape of each line. From traditional Chinese \textit{gongbi} (fine brushwork) paintings~\cite{cahill1959tao} to modern anime comic strips, diverse sketch lines create captivating artworks. When extracting sketches from the same scene, different artists produce sketches in varying styles. Textures in sketches are implied through variations in line density and stroke manipulation, with artists using textured strokes to suggest material properties and add depth~\cite{berger2013style}.

Generating sketches in styles that faithfully represent the diversity observed in manual sketches remains challenging. Traditional edge detection methods, such as Canny edge detection~\cite{canny1986computational}, Holistically-Nested Edge Detection (HED) \cite{xie2015hed}, and TEED \cite{soria2023tiny}, primarily capture uniformly thick lines and fail to emulate the artistic diversity of manual sketches. More advanced techniques, such as SketchKeras \cite{sketchkeras} and Anime2Sketch \cite{anime2sketch}, have trained models to replicate pencil strokes using paired colored images and their corresponding sketches. Recently, CLIPasso \cite{vinker2022clipasso} and CLIPascene \cite{vinker2023clipascene} have generated vector sketches but are typically limited to producing a single artistic style. 

Recent efforts aim to extract sketch styles from a data-driven perspective by learning from thousands of paired \cite{ref2sketch} and unpaired \cite{semi2sketch} datasets. These methods can reference a single image for style during inference. However, the produced styles are limited to those seen during the training phase, even when using a completely different reference style, as shown in Fig.~\ref{fig2} (a). This limitation significantly restricts their applicability in real-world scenarios, where users may wish to apply a broader range of artistic styles not represented in the training data. Consequently, the diversity and adaptability of these methods are constrained, hindering their effectiveness for users seeking to generate sketches in a wide variety of styles.

To overcome the need for extensive training and extend to arbitrary reference styles, leveraging external prior knowledge is essential. Diffusion priors are arguably the best source for obtaining diverse visual knowledge and have been employed in many zero-shot settings, particularly for style transfer techniques such as IP-Adapter \cite{ye2023ip}, StyleAligned~\cite{hertz2024style}, StyleID \cite{chung2024style}, and InstantStyle \cite{wang2024instantstyle}. However, these methods focus on overall stylistic changes and do not effectively decouple texture and contour information. Additionally, they struggle with the inherent averaging effect in color generation seen in Stable Diffusion, which tends to preserve the mean color value for natural images, resulting in washed-out appearances rather than distinct black-and-white tones, as shown in Fig.~\ref{fig2} (b). This issue is particularly problematic for sketch extraction, where high contrast and clear lines are essential.

In this paper, we propose \textbf{Mix}ture-of-\textbf{S}elf-\textbf{A}ttention (MixSA) to address the aforementioned challenges. Like diffusion-based style transfer techniques that manipulate the keys and values of cross-attention maps, our method also edits images in a similar fashion. But unlike them, our approach mimics how artists envision rough outlines before sketching, applying this concept to the initial latent space of Stable Diffusion. By replacing the keys and values of self-attention with those from reference sketches, our method seamlessly integrates brushstroke elements into the initial outline image. This mixture-of-self-attention technique effectively merges the brushstroke styles of reference sketches with the texture and contours of the original colored image.

\begin{figure*}
  \includegraphics[width=1\textwidth]{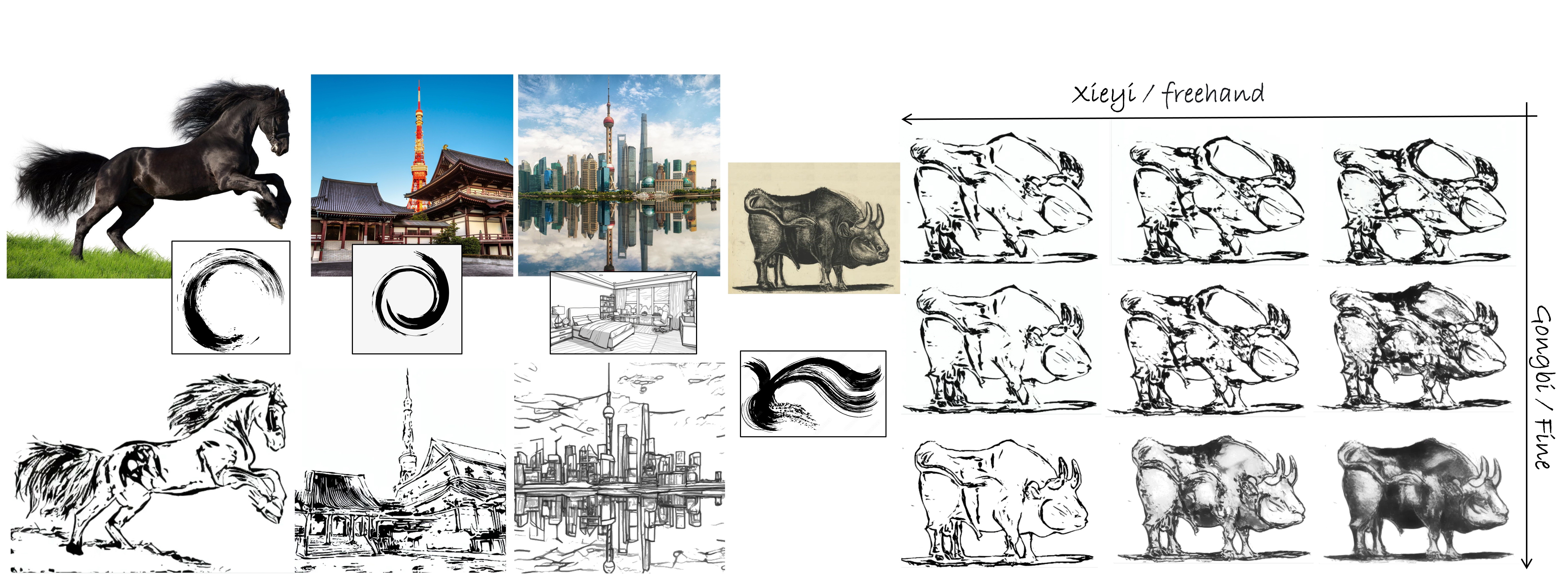}
  \caption{We propose MixSA, a training-free approach for extracting sketches from a color image using an input reference style image. Our model not only faithfully captures the input styles (left) but also allows interpolation between two styles to generate novel, unseen styles (right), exemplified by the \textit{Xieyi} (Freehand) and \textit{Gongbi} (Fine) styles.}
  \label{fig1}
\end{figure*}

We focus on replacing the self-attention keys and values in the colored image, particularly in the late decoder layers that handle local textures. This alignment ensures that the brushstroke style of the reference image harmonizes with the texture and contours of the colored image. Our method allows precise control over texture retention by adjusting the proportion of reference style's self-attention Q-values. Additionally, we transform the initial outline to match the color distribution of the final sketch, addressing the color averaging issues in Stable Diffusion. Ultimately, MixSA produces high-quality sketches that faithfully replicate the brushstroke styles of reference sketches (see Fig.~\ref{fig1}), surpassing other methods in both quality and fidelity. Moreover, our solution enables flexible application by interpolating between two styles (see the right part of Fig.~\ref{fig1}), leading to the creation of novel styles that have never been seen before. Extensive experiments on standard benchmarks across various metrics demonstrate our superiority in terms of sketch quality, flexibility, and applicability.

Our contributions can be summarized as follows:
\begin{itemize}
  \item We propose the first training-free sketch extraction method, enabling the direct use of reference style images without the need for extensive model training.
  \item We introduce the mixture-of-self-attention technique, which allows for the control of texture density and the interpolation between two different styles, enhancing the flexibility and diversity of the generated sketches.
    \item We develop a technique to address the inherent color distribution challenges in Stable Diffusion, ensuring that the generated sketches maintain the appropriate color contrasts and details reflective of the reference sketches.
\end{itemize}

\begin{figure}[t]
  \centering
  \includegraphics[width=\linewidth]{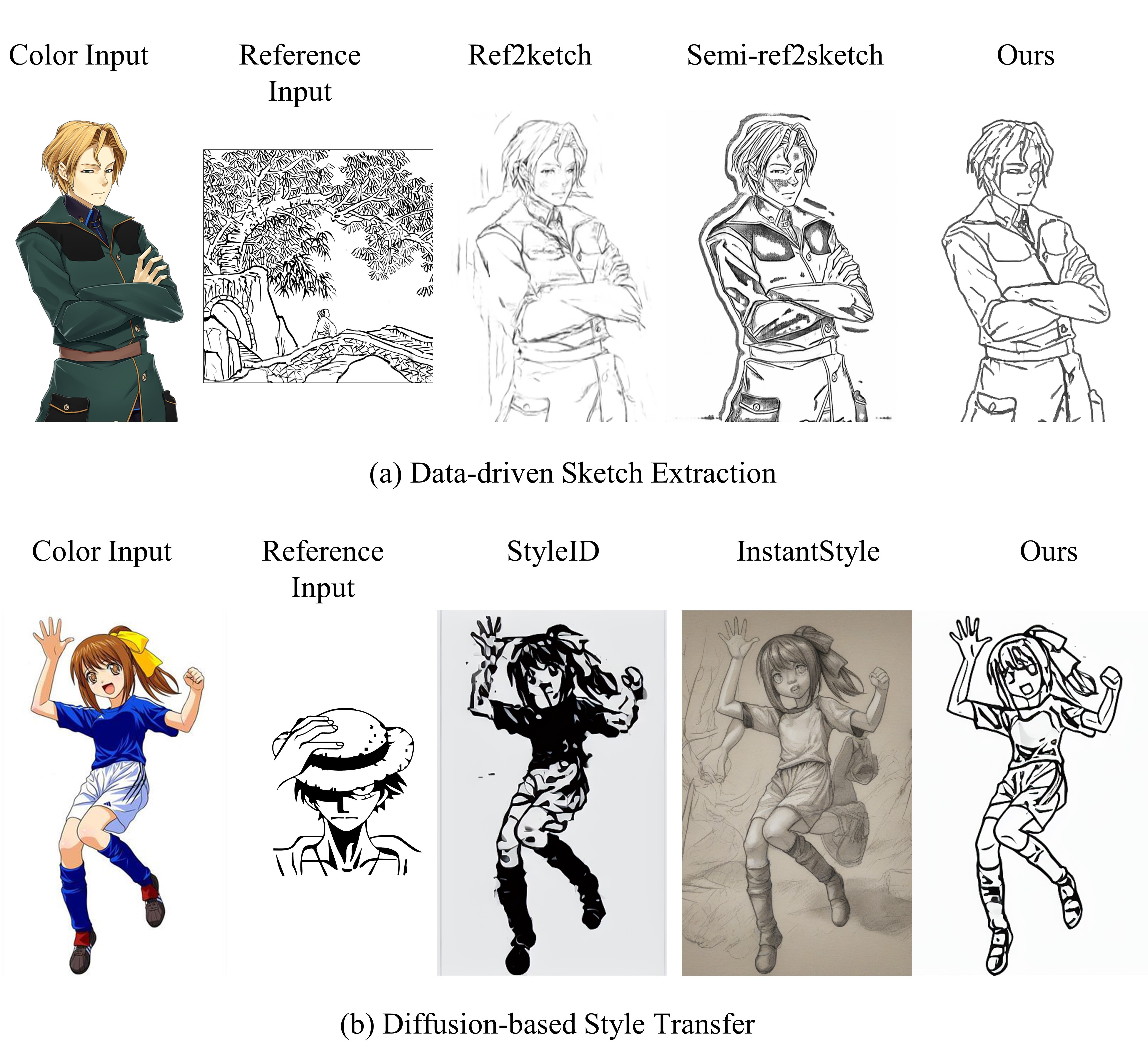}
  \vspace{-2mm}\caption{Data-driven sketch extraction methods (a) struggle to adapt to unseen reference styles, while diffusion-based style transfer methods (b) fail to disentangle overall styles from sketch-specific styles, resulting in inconsistent sketch transfers. Our MixSA overcomes both the extensive training requirements and the sketch style transfer limitations.}\vspace{-3mm}
    \label{fig2}
\end{figure}

\section{Related work}
\subsection{Sketch Extraction}
Sketch extraction, an advanced form of edge detection, has evolved significantly from traditional methods to deep learning-based techniques. Initially, edge detection focused on identifying abrupt changes in image intensity, with seminal works like Canny's edge detector~\cite{canny1986computational} providing robust techniques for capturing edge features. However, these early methods were limited in rendering artistic sketches.

The integration of deep learning transformed edge detection, leading to sophisticated models capable of capturing a broader range of nuances suitable for sketches. Notable advancements include Holistically-Nested Edge Detection (HED)~\cite{xie2015hed}, which leveraged deep neural networks to learn rich hierarchical representations, and the Bi-Directional Cascade Network (BDCN)~\cite{he2019bi}, which refined edge detection by focusing on perceptual quality. More recent methods like TEED~\cite{soria2023tiny}, UAED~\cite{zhou2023treasure}, and DexiNed~\cite{poma2020dense} have continued to improve precision and detail in edge detection.

While effective for edges, these methods often lacked the subtlety required for full sketch extraction, which also captures the artistic style and texture of sketch lines. The field then transitioned to models that could interpret artistic nuances, moving from pure edge detection to sketch extraction. Approaches like SketchKeras~\cite{sketchkeras} and Anime2Sketch~\cite{anime2sketch} addressed these needs by mimicking artistic strokes, and APDrawingGAN~\cite{YiLLR19} used hierarchical GANs to generate artistic portrait drawings from face photos. Additionally, several methods have been specifically developed for face sketch synthesis~\cite{zhang2020npgm,zhang2019dtfss,zhang2019dllfss,8764602}, focusing on preserving identity-specific features and common facial structures to improve the accuracy of sketch generation. These advancements have significantly enhanced the ability to handle face sketches across various contexts.

However, these models depended on paired datasets, limiting their flexibility. This challenge was partly addressed by methods like CLIPasso~\cite{vinker2022clipasso}, which adapted to various artistic styles without direct style pairing, indicating a significant shift towards more versatile sketch extraction methodologies.

Recent techniques emphasize learning from unpaired data or single reference sketches, crucial for reducing the barrier to entry for personalized sketch extraction models. However, learning from a single sketch image introduces challenges in style consistency and transferability. Methods like ref2sketch~\cite{ref2sketch} and semi-ref2sketch~\cite{semi2sketch} cluster sketches by style to improve training efficiency and output consistency, but often struggle with the diversity and subtlety of artistic styles, particularly with unseen references.

Our work builds on these advancements by introducing MixSA, a flexible, style-adaptive sketch extraction model leveraging latent diffusion models and DDIM inversion techniques. This approach maintains high fidelity to reference styles, allowing for real-time adaptation and high-quality sketch generation without any training.

\subsection{Style Transfer}

In the broader context of image-to-image translation, style transfer adapts the style of one image to the content of another~\cite{zhang2023unified}. Gatys et al. \cite{gatys2016image} introduced a neural algorithm for artistic style transfer using convolutional neural networks (CNNs) to separate and recombine content and style from images. While this method achieves impressive texture transfer results, it often fails to preserve the fine details necessary for sketch extraction. Huang and Belongie~\cite{huang2017arbitrary} proposed Adaptive Instance Normalization (AdaIN) for real-time style transfer by aligning the mean and variance of content feature maps with those of style feature maps. Traditional style transfer methods focus on transferring textures and colors to create varied artistic effects. However, for sketches, the task requires not only style adaptation but also the preservation of line and form.

The advent of diffusion models (DMs)~\cite{rombach2022high,liu2023painterly, zhang2023inversion, yang2023zero, ma2023rast} has revolutionized the field by providing a framework for high-fidelity style adaptation. DMs generate images from a noise distribution without requiring paired examples, making them well-suited for high-quality style adaptation tasks. Recent advances in image-to-image translation, such as Prompt-to-Prompt~\cite{mokady2023null}, Ip-Adapter~\cite{ye2023ip}, FateZero \cite{qi2023fatezero}, Stylediffusion~\cite{wang2023stylediffusion}, and ELITE \cite{wei2023elite}, enhance high quality image generation capabilities by manipulating text conditions for cross-attention while maintaining the underlying attention map.

Innovative techniques such as InST~\cite{zhang2023inversion}, Cross-Image-Attention~\cite{alaluf2023cross}, StyleAligned~\cite{hertz2024style}, Swapping Self-Attention~\cite{jeong2024visual}, StyleID~\cite{chung2024style}, and InstantStyle~\cite{wang2024instantstyle} adapt diffusion models for direct style transfer. These approaches harness the latent encoding capabilities of diffusion models and incorporate mechanisms for dynamic style adaptation from reference images without the need for retraining. This allows for the creation of style-consistent images using a reference style through a straightforward inversion operation.

The essence of style transfer involves transforming textural elements from a reference image, which differs from the needs of sketch extraction that focuses on line and form. Traditional style transfer models often fall short for sketches, where maintaining the integrity of edges and lines is crucial. Our work addresses these limitations using diffusion models and DDIM inversion to ensure high fidelity in line and form preservation, creating sketches that maintain the essential characteristics of the reference style and providing a robust solution for artistic sketch rendering.

\section{Preliminaries}

In this section, we provide a detailed overview of the fundamental concepts and techniques used in our approach, including Latent Diffusion Models, DDIM Inversion, and the Attention Mechanism in Stable Diffusion.

\begin{figure*}[h]
  \includegraphics[width=\linewidth]{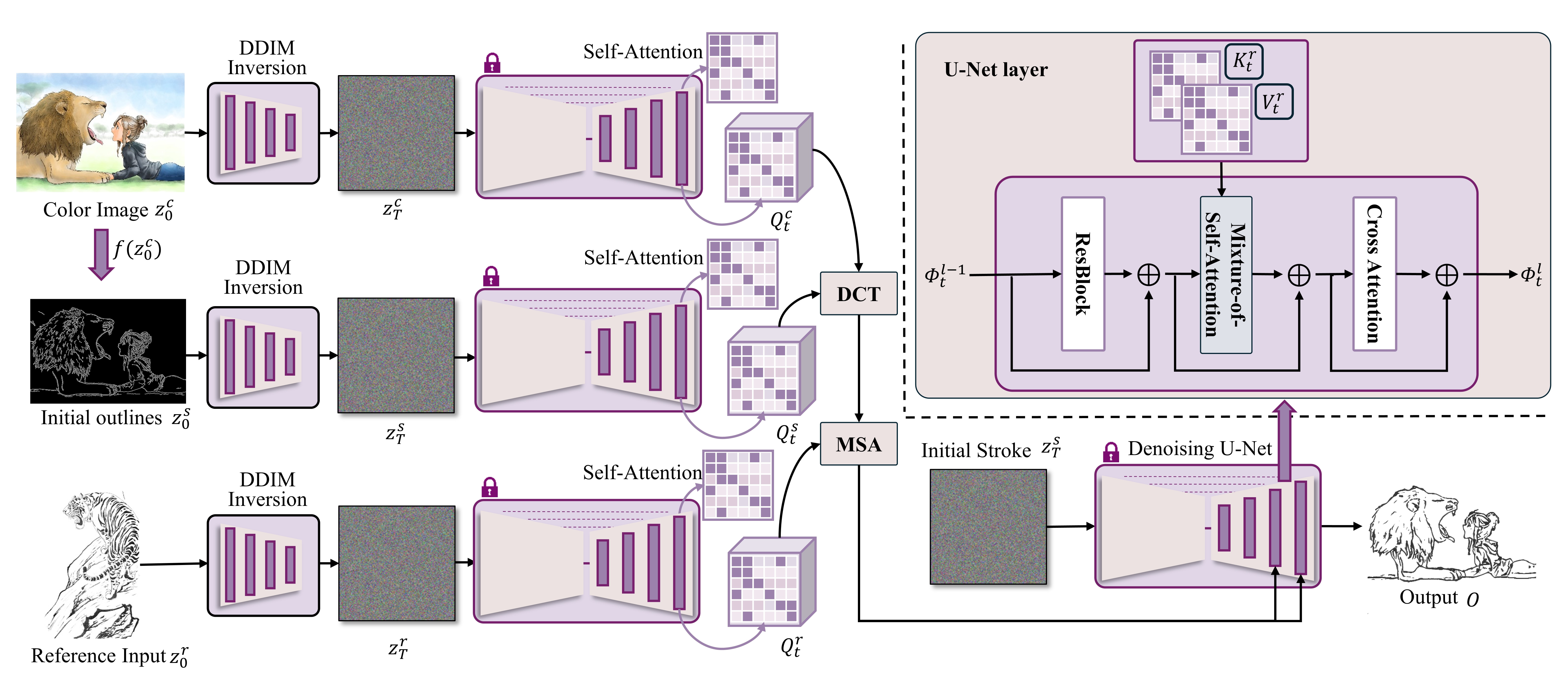}
  \vspace{-5mm}\caption{The architecture of our proposed MixSA begins with a color image \( z^c_0 \) and its initial outlines \( z^s_0 \), both converted to latent representations via DDIM inversion. A reference sketch \( z^r_0 \) undergoes the same processing. The self-attention features (\( Q_t \)) from these latent representations are manipulated using the mixture-of-self-attention (MSA) and Decomposing the Contours and Texture (DCT) modules. These modified features are injected into the denoising U-Net, integrating the reference sketch's key and value features (\( K_t^r \) and \( V_t^r \)) into the decoder. The final output \( O \) is a high-fidelity sketch that faithfully matches the reference style.}
  \label{fig3}
\end{figure*}

\subsection{Latent Diffusion Models}
Diffusion models are powerful generative models that synthesize data samples from Gaussian noise through a series of iterative denoising steps \cite{rombach2022high}. Our approach leverages a state-of-the-art text-conditioned Latent Diffusion Model (LDM), Stable Diffusion (SD), which operates in the latent space of a pre-trained image autoencoder rather than directly in the image space.

In this model, an image \( x \) is first encoded into latent representations \( z \) by a pre-trained autoencoder. This autoencoder compresses the high-dimensional image \( x \) into a lower-dimensional latent space representation \( z \), capturing essential image features. The denoising process is handled by a time-conditional U-Net~\cite{ronneberger2015u} \( \epsilon_{\theta} \), which iteratively refines these latent representations.

Initially, a random noise vector \( z_T \sim N(0, 1) \) is sampled and progressively denoised to obtain the latent representation \( z_0 \), which is then decoded back into an image. This process can be mathematically described as:
\begin{equation}
\label{eq1}
z_{t-1} = \frac{z_t - \sqrt{1-\beta_t} \epsilon_{\theta}(z_t, t)}{\sqrt{1 - \beta_{t-1}}} + \sigma_t \epsilon,
\end{equation}
where \( \beta_t \) represents the noise schedule, \( \sigma_t \) is a noise scaling factor, and \( \epsilon \) is random Gaussian noise. The denoising steps gradually remove the noise, reconstructing the image from the latent representation \cite{rombach2022high, ronneberger2015u}.

\subsection{DDIM Inversion}
DDIM (Denoising Diffusion Implicit Models) inversion allows precise control over the generation process by inverting the diffusion from a learned latent representation back to the data space. This technique is crucial for tasks requiring direct manipulation of image features in their latent form.

In our sketch extraction approach, DDIM inversion manipulates the latent space representation of color images and sketches. Given a starting image \( x_0 \) and its noisy version \( x_t \), the inversion process estimates \( x_{t-1} \) from \( x_t \) using the formula:
\begin{equation}
\label{eq2}
x_{t-1} = \sqrt{\bar{\alpha}_{t-1}} \left( \frac{x_t - \sqrt{1-\bar{\alpha}_t} \epsilon_\theta(x_t, t)}{\sqrt{\bar{\alpha}_t}} \right) + \sigma_t \epsilon,
\end{equation}
where \( \bar{\alpha}_t \) is the cumulative product of noise levels at each timestep. This reverse process is iteratively applied until the original image is reconstructed or transformed into a target state.

By leveraging DDIM inversion, we can guide the model to retain or enhance specific features like edges and textures, crucial for high-fidelity sketch representations. This allows for additive and subtractive operations in the latent space, aligning closely with artistic goals and ensuring the integrity of edge details while subtly altering textures to match the sketch style.

\subsection{Attention Mechanism in Stable Diffusion}
The denoising U-Net \( \epsilon_{\theta} \) within the SD model incorporates a series of basic blocks, each comprising a residual block, a self-attention module, and a cross-attention module. During a denoising step \( t \), features from the previous layer \( (l-1) \) pass through the residual block to generate intermediate features \( f^l_{t} \).

The self-attention layer reorganizes these features, facilitating long-range interactions between image elements. This is particularly important for capturing complex dependencies within the image. The self-attention mechanism operates by projecting the features into queries \( Q \), keys \( K \), and values \( V \) and computing the attention as follows:
\begin{equation}
\label{eq3}
\hat{f}^l_{t} = \text{Softmax}\left(\frac{QK^T}{\sqrt{d}}\right)V,
\end{equation}
where \( d \) denotes the dimension of the projected query \cite{vaswani2017attention}.

The cross-attention layer, typically used for integrating textual information from a text prompt \( P \), is bypassed in the implementation due to the focus on transferring referenced stroke styles without text conditions. This setup allows us to inject the style of strokes from reference sketches directly into the denoising process, facilitating sketch extraction and faithful image synthesis \cite{rombach2022high, ronneberger2015u}.

\section{Method}\label{method}

Our goal is to extract sketches \(O\) from given colored images \(z^c_0\) while mimicking the style of a reference sketch \(z^r_0\). Due to the scarcity of paired data between sketches and colored images, creating large-scale paired datasets is resource-intensive and limited in capturing stylistic diversity. To address this, our approach leverages the self-attention mechanism to extract sketch line features from a given reference image and apply them to the sketch generation process of the colored image.

Similar to previous works \cite{semi2sketch} that used edge detection networks for line loss, we utilize a variety of edge detection algorithms (TEED~\cite{soria2023tiny} is used by default due to its robustness) for Contour Detection (CD) in colored image as the initial sketch \(z^s_0\). We then obtain the latent space representations of the colored image \(z^c_T\), edge image \(z^s_T\), and reference sketch image \(z^r_T\) through DDIM inversion, which assumes that features of the colored image and edge image in the latent space can be added and subtracted. During the sketch generation process, we inject the key and value pairs (\(K\) and \(V\)) of the reference sketch image into the latter layers of the decoder, which are related to line generation, through shared attention operations. This method allows for a nuanced and style-accurate rendering of sketches, integrating the distinctive line styles and textures of the reference sketch into the final output. See Fig.~\ref{fig3} for an overview of our network design.

\subsection{Mixture of Self-Attention}  
Inspired by recent style transfer techniques with diffusion models, such as Swapping Self-Attention~\cite{jeong2024visual}, StyleAligned~\cite{hertz2024style}, and StyleID~\cite{chung2024style}, we extend these concepts into the domain of self-attention for sketch extraction. Our method focuses on transferring the textural attributes of a reference image into a sketch by substituting the key (\(K\)) and value (\(V\)) pairs of the colored image with those derived from the style of the reference image during the generation process.

This process begins by inverting the representations of the reference image, colored image, and edge image from their image state (\(t = 0\)) to Gaussian noise (\(t = T\)) through DDIM inversion. During this inversion, we capture the self-attention features (\(K^r_t, V^r_t\)) of the reference image across predefined timesteps \(t = 0, \ldots, T\). A naive approach would be to apply \(K^r_t\) and \(V^r_t\) directly to the denoising process of the colored image using its latent space and query features. However, this does not work effectively for sketch extraction.

\begin{figure}[t]
  \includegraphics[width=\linewidth]{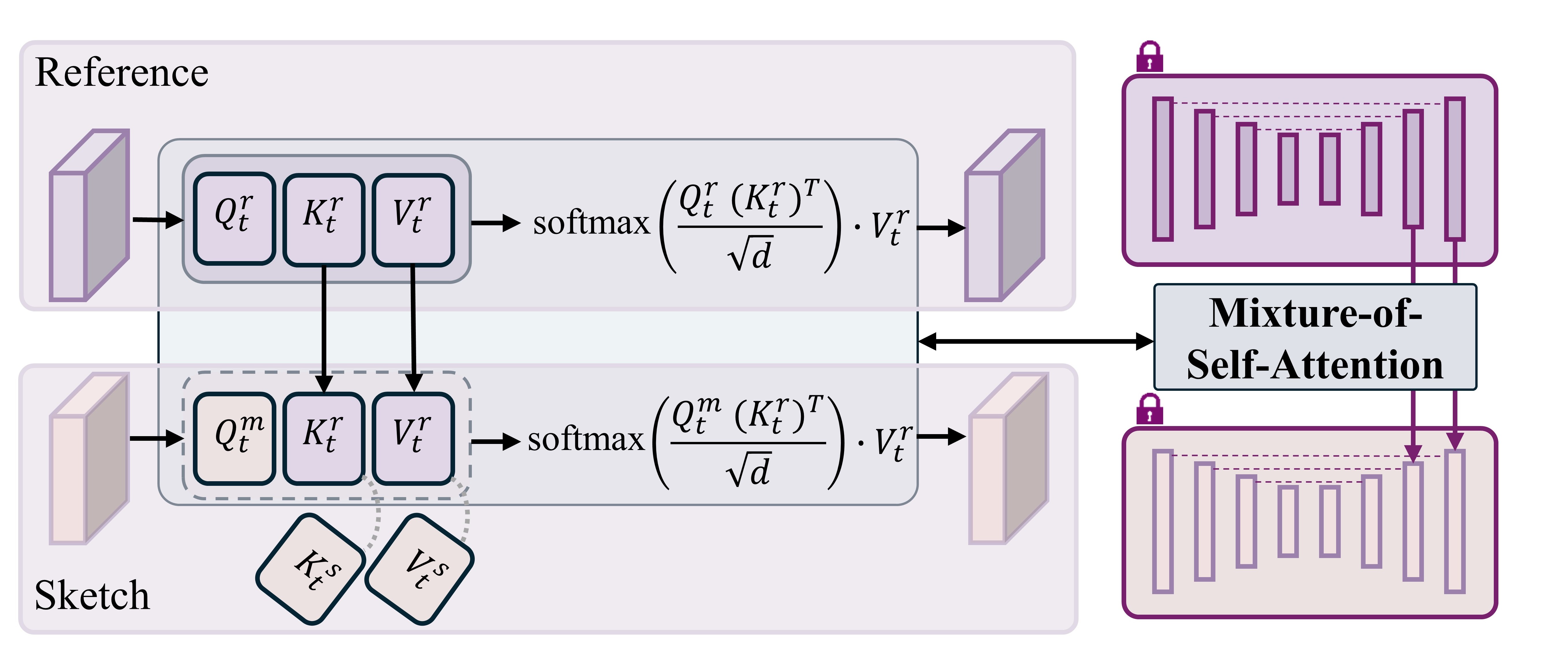}
\caption{Detailed mechanism of the mixture-of-self-attention module. The reference image's self-attention features (\( Q_t^r, K_t^r, V_t^r \)) and the initial sketch's self-attention features (\( Q_t^m \)) are computed, with \( Q_t^m \) being a fusion of the color image and its initial outlines. The reference's key and value features (\( K_t^r \) and \( V_t^r \)) replace those of the sketch in the self-attention mechanism. This ensures that the generated sketch incorporates the brushstroke styles of the reference image while maintaining the structure and layout of the initial outlines.}
  \label{fig4}
\end{figure}

Our strategy involves using the edge image as the initial latent noise for sketch generation, denoted as \(z^{s}_T\). This approach not only helps mitigate color leakage from the colored image during sketch generation but also retains the texture and edge features of the colored image for weighted coupling of texture and edges (detailed in Section 3 of this chapter). We replace the original \(K^{s}_t, V^{s}_t\) of the edge image with the features \(K^r_t, V^r_t\) from the reference style image and use a weighted mixture of query features from the colored and edge images, \(Q^m_t\), as the query for calculating attention in the sketch generation process, as illustrated in Fig.~\ref{fig4}. This integration is articulated through the following operation:
\begin{equation}  
\label{eq4}
\phi ^{s}_{out} = \text{Attn} (Q^m_t, K^r_t, V^r_t),  
\end{equation}
where \(Q^m_t\) represents the fused query features for the sketch. The detailed computation of this fused query feature will be elaborated in the next two subsections. This method ensures that the generated sketch incorporates the brushstroke styles of the reference image while maintaining the structure and layout of the initial outlines.

\subsection{Decomposing Contours and Texture (DCT) Module}

Preserving object textures is essential for different sketch styles, making it crucial to separate textures from edge contours to allow controllable manipulation. Notably, artists often begin sketch creation by drawing initial contours before refining details. Inspired by this process, we use edge detection algorithms to create the initial sketch. We provide a selection of edge detection options to generate these initial contours. 

We assume that features of the colored image can be additively and subtractively manipulated in the latent space. Based on this assumption, we use a parameter, \( \beta \) (ranging from 0 to 1), to control the mixture of the colored image and the sketch image features in the sketch generation process. This is formulated as follows:
\begin{equation}
\label{eq5}
Q^{cs}_t = \beta \times Q^c_t + (1 - \beta) \times Q^{s}_t
\end{equation}

Here, \( Q^{cs}_t \) represents the mixed query feature at time step \( t \), \( Q^c_t \) is the feature from the colored image, and \( Q^s_t \) is the feature from the edge image. The parameter \( \beta \) balances the influence of edge details and textural elements, allowing for a nuanced integration of texture and line work. During our experiments, we found that different layers of the UNet focus on generating different attributes of the image. For sketches, we discovered that injecting features at the 10th and 11th layers yielded the best results.

This method enables us to blend the characteristics of the colored image and the sketch style image in the latent space, achieving a precise replication of the artistic intent of the reference sketch while maintaining high fidelity to the original textures and contours.

\subsection{Sketch Style Interpolation}

\subsubsection{Target Style 1}

As demonstrated in Fig.~\ref{fig1}, control over the degree of freehand expression is achieved through two axes, allowing for varying levels of abstraction as shown in a single row of our abstract matrix. Before injecting the reference style brush strokes, we calculate an intermediary representation \(Q^{cs}_t\) using a weighted combination of \(Q^c_t\) and \(Q^s_t\). We introduce a parameter \(\zeta\) to adjust the proportion of the style injected, as described by equation~\ref{eq6}. A higher \(\zeta\) value results in greater preservation of \(Q^r_t\), making the sketch strokes closer to the reference image's brush strokes. Conversely, a lower \(\zeta\) value means more of \(Q^cs_t\) is injected, bringing the sketch strokes closer to the original texture and contour, thus presenting a more freehand effect artistically.

\begin{equation}
\label{eq6}
Q^{m}_t = \zeta \times Q^{cs}_t + (1 - \zeta) \times Q^{r}_t,
\end{equation}

where \(\zeta\) ranges from 0 to 1. These operations are specifically applied in the latter stages of the decoding process (typically within the 10th-11th layers of a standard decoder), which are crucial for detailing local stroke textures and nuances.

\begin{figure}[t]
  \includegraphics[width=\linewidth]{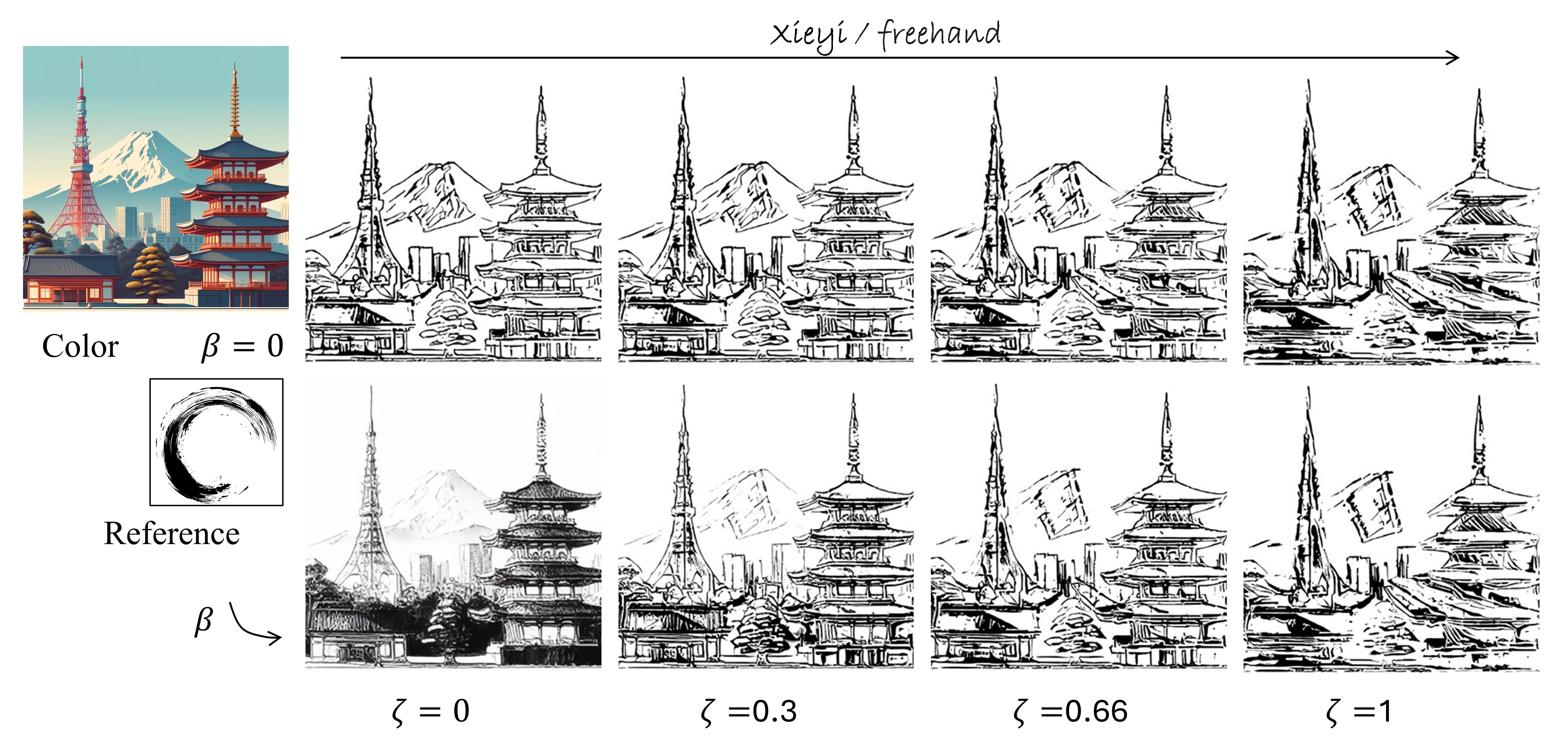}
 \caption{Illustration of \(\zeta\) as the control axis for the degree of freehand (Xieyi) style. In the first row, with \(\beta = 0\) (no texture), the scene's sketch is automatically extracted based on object contours. From left to right, the sketch becomes increasingly freehand as \(\zeta\) increases, with brush stroke styles approaching those of the reference sketch. In the second row, \(\beta\) decreases from left to right, reducing texture inclusion. Both object contours and textures influence the sketch extraction. As the degree of freehand style increases, the sketch transitions from detailed to more abstract representations.}
  \label{fig5}
\end{figure}

As shown in Fig.~\ref{fig5}, this enables flexible control over the degree of style transfer, where a higher \(\zeta\) preserves more of the original texture, and a lower \(\zeta\) amplifies the impact of the reference style’s stroke characteristics. This capability enhances the artistic fidelity of the generated sketches and allows for a customizable balance between the influence of the original colored image and the artistic style introduced by the reference sketch. By employing this refined approach to style transfer in sketch extraction, we ensure that the resulting sketches maintain a harmonious blend of original and stylized features.

\subsubsection{Target Style 2}

Inspired by the Semi-ref2sketch~\cite{semi2sketch}, we observed that varying degrees of texture preservation from the original colored image are crucial for representing different sketch styles. To simulate the presentation from sparse to detailed textures, we introduce \(\beta\) in Equation~\ref{eq9} to implement different weights for texture and contour proportions. Through the \(\beta\) parameter, the model adjusts the degree of texture retention during sketch rendering, displaying different sketch results.

\begin{figure}[t]
  \includegraphics[width=\linewidth]{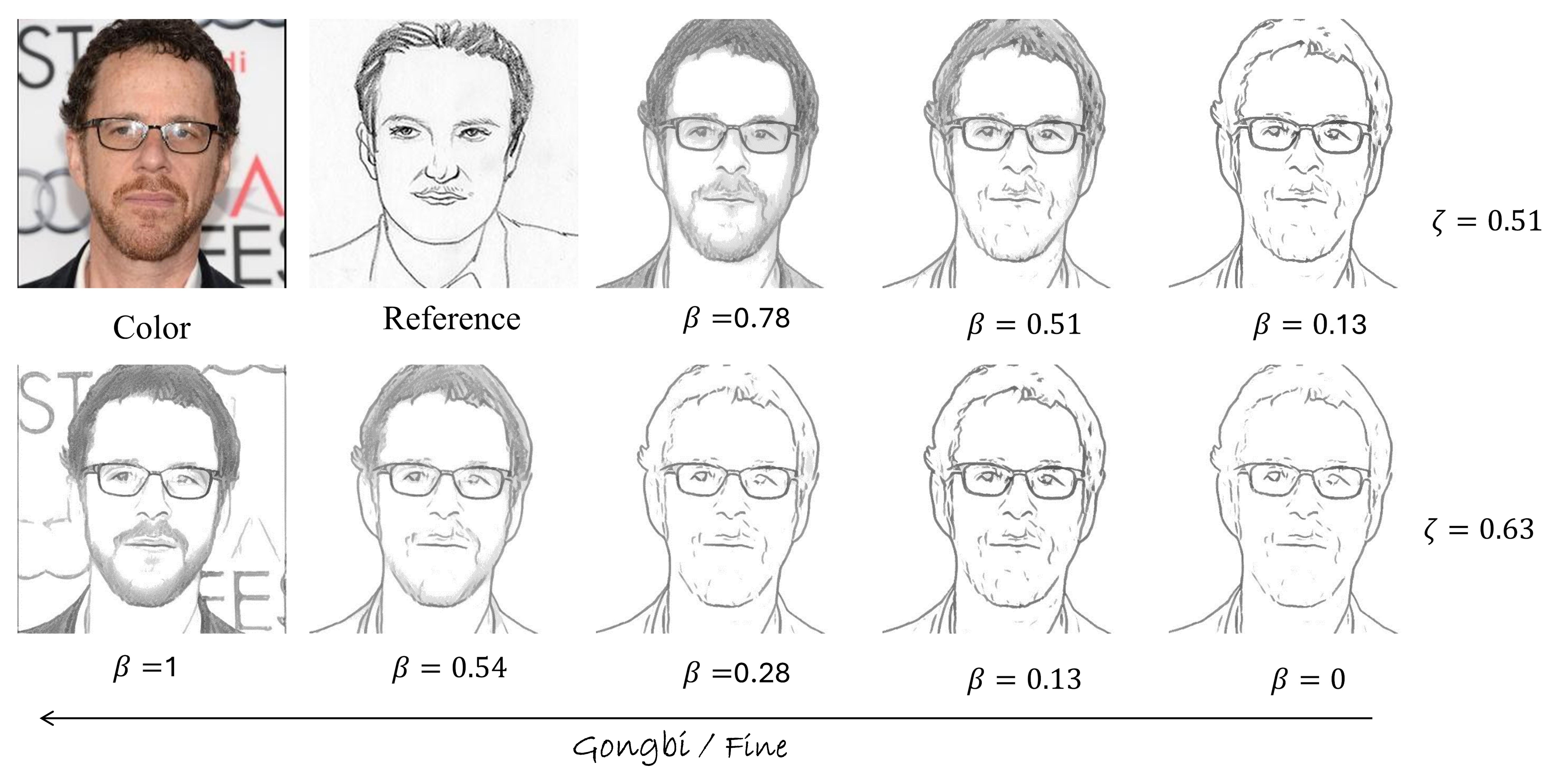}
\caption{Illustration of \(\beta\) as the control axis for the degree of fine (\textit{Gongbi}) style. In the first row, with a fixed \(\zeta\), the sketches become increasingly detailed and textured as \(\beta\) increases. In the second row, the first image on the left shows the sketch result without foreground extraction by U2-Net. Subsequent images demonstrate that, with a fixed \(\zeta\), increasing \(\beta\) enhances the interplay between object contours and textures, making the sketch progressively closer to the grayscale representation of the original image.}
  \label{fig6}
\end{figure}

As shown in Fig.~\ref{fig6}, the \(\beta\) parameter allows for varying degrees of realism or fine brushwork effects. With a fixed \(\zeta\), increasing \(\beta\) enhances the interplay between object contours and textures, making the sketch progressively closer to the grayscale representation of the original image. This achieves varying degrees of detail and realism, providing a versatile tool for generating sketches with different stylistic nuances. 

\subsubsection{More Flexible Sketch Extraction}
By manipulating two parameters, \(\zeta\) and \(\beta\), we can generate sketches with varying degrees of fidelity to the reference sketch image, texture retention, and sketch sparsity. Specifically:

1. The parameter \(\zeta\) determines the degree to which the generated sketch adheres to the reference sketch image. A higher \(\zeta\) value enhances the preservation of the reference's brushstroke styles, resulting in a sketch more similar to the reference. Conversely, a lower \(\zeta\) value retains more features from the colored image, producing a sketch closer to the original colored image.

2. The parameter \(\beta\) controls the inclusion of textures in the sketch. A higher \(\beta\) value incorporates more textural elements from the colored image, enriching the sketch with details. A lower \(\beta\) value reduces the inclusion of textures, emphasizing the contour features.

3. Parameter Adjustment Challenges. Adjusting $\zeta$ and $\beta$ provides flexibility but also presents certain challenges. When $\beta = 0$, the sketch is generated based solely on edge information, lacking texture details. Increasing $\beta$ introduces more texture from the original image (see Fig. 15). The $\zeta$ parameter controls the alignment with the reference style—setting $\zeta$ too high may result in content leakage, while setting it too low may prevent proper alignment. For further details, please refer to the supplementary materials.

4. We offer three edge detection methods for initial stroke extraction: Canny~\cite{canny1986computational}, TEED~\cite{soria2023tiny}, and HED~\cite{xie2015hed}. TEED is the default, but users can select the method that best suits their needs.

By appropriately setting these parameters, the model can generate sketches that range from very sparse and abstract to highly detailed and realistic.

Substituting Eq.~\ref{eq5} and Eq.~\ref{eq6} into Eq.~\ref{eq4} yields the following equation for \(\phi ^{s}_{out}\):

\begin{equation}
\label{eq7}
\phi ^{s}_{out} =\text{Attn} ( \zeta \times (\beta \times Q^c_t + (1 - \beta) \times Q^{s}_t) + (1 - \zeta) \times Q^{r}_t, K^r_t, V^r_t)
\end{equation}

\begin{figure}[t]
  \includegraphics[width=\linewidth]{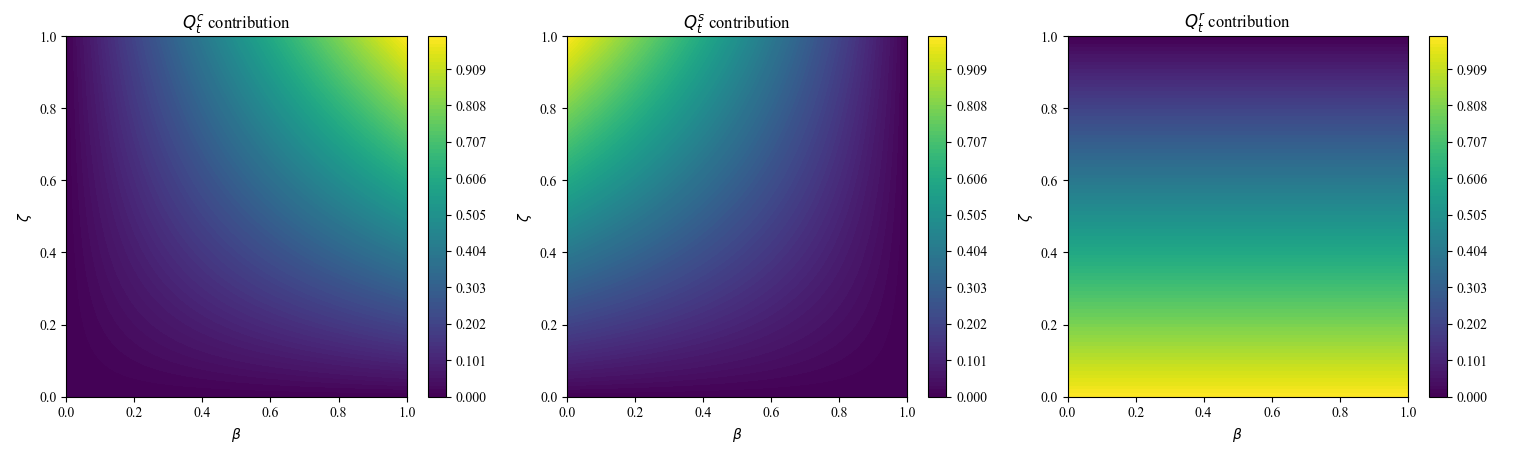}
  \caption{Visualization of the contributions of \(Q^c_t\), \(Q^s_t\), and \(Q^r_t\) to \(\phi ^{s}_{out}\) under varying \(\zeta\) and \(\beta\) values. The plots illustrate how different parameter settings influence the final output, with color gradients representing the magnitude of each contribution.}
  \label{fig7}
\end{figure}

In this context, \(\zeta\) and \(\beta\) serve as flexible control parameters that enable precise adjustments to the sketch generation process, balancing between the influence of the reference image and the original colored image. A more intuitive presentation is detailed in Figure~\ref{fig7}. For a more detailed analysis of the trade-off between texture retention and reference adherence, controlled by the parameters \(\zeta\) and \(\beta\), please refer to the supplementary material, where further discussion and examples are provided (see Sec. A\uppercase\expandafter{\romannumeral1} and Fig. A2 in the supplementary material.

\subsection{Color Averaging Artifact of Stable Diffusion}

When utilizing diffusion models like Stable Diffusion (SD) for sketch extraction, a significant challenge is the inherent averaging effect in color generation. This becomes problematic when converting color images to sketches, where high-contrast black-and-white imagery is essential. The diffusion process tends to preserve the mean color value around 0.5, leading to a washed-out appearance instead of distinct black and white tones.

The transformation of an image into noise involves adding Gaussian noise over a series of steps. Consider an image \( x_0 \) which undergoes noise addition to become a noisy image \( x_T \). This process is described by the iterative equation:
\begin{equation}
\label{eq8}
x_t = \sqrt{\bar{\alpha}_t} x_0 + \sqrt{1 - \bar{\alpha}_t} \epsilon,
\end{equation}
where \( \epsilon \sim \mathcal{N}(0, I) \) represents the added Gaussian noise, and \( \bar{\alpha}_t \) is the cumulative product of noise scalars over \( t \) steps.

\subsubsection{High-Frequency vs. Low-Frequency Noise}

High-frequency components correspond to fine details in the image, such as edges and textures, which have smaller spatial correlations and are more sensitive to noise. Low-frequency components represent broader, smoother areas of the image with larger spatial correlations.

To fully disrupt these components, consider the power spectral density (PSD) of the image. High-frequency components dominate at large frequencies, while low-frequency components dominate at lower frequencies. The noise addition process can be analyzed in the frequency domain, where higher frequencies are attenuated more rapidly due to noise.

The variance of noise added at each step is given by \( \sigma^2_t = 1 - \bar{\alpha}_t \). For a high-frequency component \( f_H \) and a low-frequency component \( f_L \), the steps required to disrupt these components can be expressed as:

\begin{equation}
\label{eq9}
\sum_{t=1}^T \text{Var}(f_H) \gg \sum_{t=1}^T \text{Var}(f_L).
\end{equation}

This inequality indicates that high-frequency components reach a complete noise state faster than low-frequency components because the cumulative variance of the noise grows more rapidly for high-frequency components. This is crucial for sketch extraction, as it emphasizes the need for techniques that preserve high-contrast details while managing the averaging tendencies of the diffusion process.

\subsubsection{Reconstruction Difficulty}

The reconstruction process in diffusion models involves reversing the noise addition to recover the original image. The challenge lies in accurately predicting and removing the noise, particularly for high-frequency components. The accuracy of reconstructing a component is inversely related to the noise variance accumulated over the diffusion steps. Since high-frequency components accumulate more noise variance, they are harder to reconstruct. Mathematically, the reconstruction error can be represented as:

\begin{equation}
\label{eq10}
\text{Error}(f_H) \propto \sum_{t=1}^T \text{Var}(f_H) \gg \text{Error}(f_L) \propto \sum_{t=1}^T \text{Var}(f_L).
\end{equation}

Thus, the accumulated noise variance for high-frequency components results in a higher reconstruction error compared to low-frequency components, making the former harder to recover.

Our goal is to generate sketches where the final output closely resembles the initial contour map of the original colored image in both color and structure. To achieve this, we use the initial contour map \( z^s_0 \) of the colored image, encoded through the diffusion model to obtain \( z^s_T \), as the initial noise for denoising in the sketch generation process. This method mitigates the issue of SD generating average-colored images by anchoring the sketch generation to a contour map aligned with the original image.

However, when the reference sketch features lighter colors, the resulting sketch may still appear overly gray due to the diffusion model's tendency to average pixel values. Mathematically, this can be represented as:

\begin{equation}
\label{eq11}
x_{t-1} \approx \frac{x_t}{2} + \frac{\epsilon}{2},
\end{equation}

where the averaging effect reduces the contrast necessary for sharp sketch lines. To address this, we introduce a binarization post-processing step to the output sketch, enhancing the contrast and making the sketch lines more prominent.

\begin{figure}[t]
  \centering
  \subfloat[]{%
    \begin{minipage}[b]{0.49\linewidth}
      \centering
      \includegraphics[width=\linewidth]{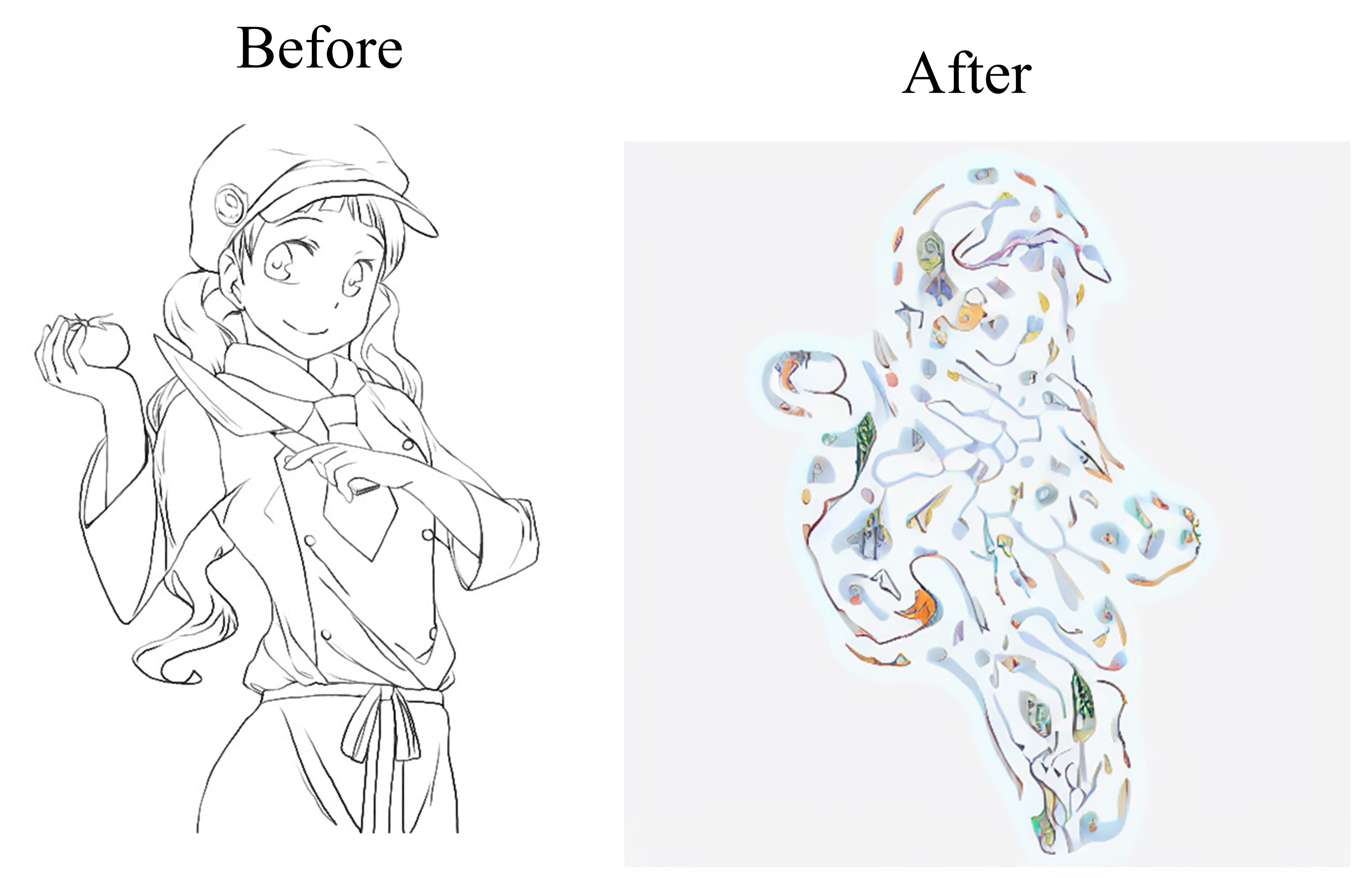}
    \end{minipage}
  }
  \subfloat[]{%
    \begin{minipage}[b]{0.49\linewidth}
      \centering
      \includegraphics[width=\linewidth]{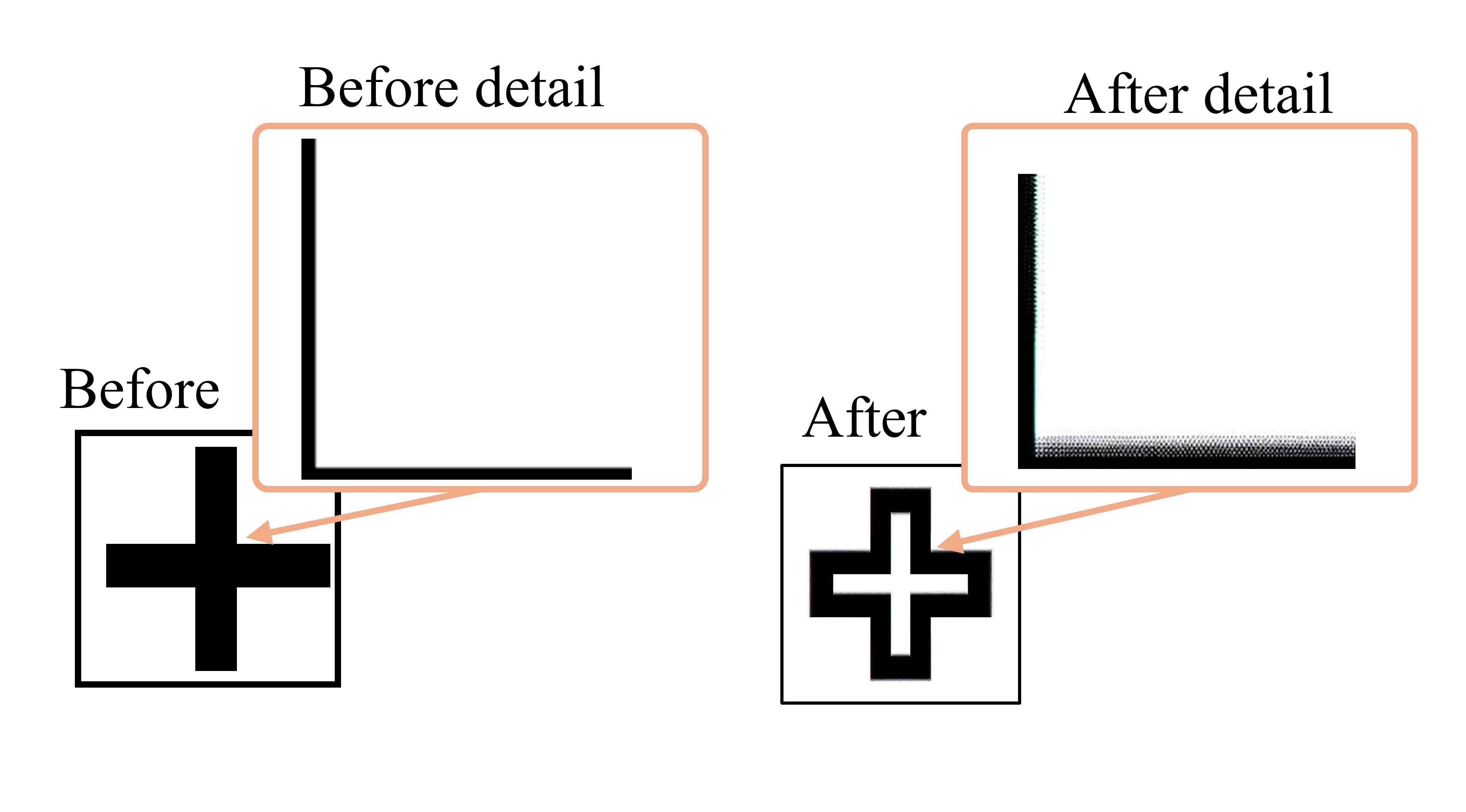}
    \end{minipage}
  }
  \caption{Illustration of the challenges faced by existing diffusion models in reconstructing sketch images characterized by extreme value distributions. (a) The original sketch input and its corresponding reconstruction after encoding and decoding through the diffusion model (step = 50, without guided conditions), highlighting issues with color reconstruction. (b) The binary sketch input with pronounced edges and its corresponding reconstruction after extending the steps demonstrate significant degradation of edge information.}
  \label{fig8}
\end{figure}

As illustrated in Fig.~\ref{fig8}, the left image is a sketch from the 4SKST dataset \cite{semi2sketch}. After processing by the SD model, the reconstructed image shows a loss of sharpness in the sketch lines. Similarly, the right image is an L-shaped binary sketch, with its reconstructed counterpart showing similar degradation.

Research has identified the averaging issue as a consequence of Gaussian noise's effect on the image's pixel distribution during the diffusion process~\cite{si2023freeu}. To mitigate this, some studies~\cite{diffusion_offset_noise} have explored fine-tuning SD models against modified noise profiles or targeted image characteristics. For example, fine-tuning against a solid black image has been tested, but even extensive iterations fail to overcome the model's propensity to revert to mid-tone averages.

To address the issue of color averaging, we apply a binarization step to the output sketch, enhancing contrast and making lines more prominent. Using a model checkpoint fine-tuned on black-and-white images~\cite{diffusion_offset_noise}, we improve the preservation of sharpness in sketch generation. However, near-binary sketches may still produce gray shadows due to zero-valued pixels affecting the denoising process in Stable Diffusion. To counter this, we binarize pixels with extreme brightness (values above 230).

To minimize the negative impact on metrics like FID and SSIM, we use adaptive smoothing and contrast enhancement. Bilateral filtering is employed to reduce noise while preserving edge details, resulting in improved sketch quality and lower FID scores. 

\subsection{Scene Decomposition}

The previous procedure considers the entire scene as a single entity. However, in practice, it is often more desirable to focus on sketching the foreground subjects while omitting the background, which can be distracting or irrelevant to the user's intention. To address this, we integrate a pre-trained U2-Net~\cite{qin2020u2} to effectively separate and extract prominent foreground objects. This allows users to selectively apply the sketch extraction process to the desired parts of the image.

\begin{figure}[t]
  \centering
  \includegraphics[width=\linewidth]{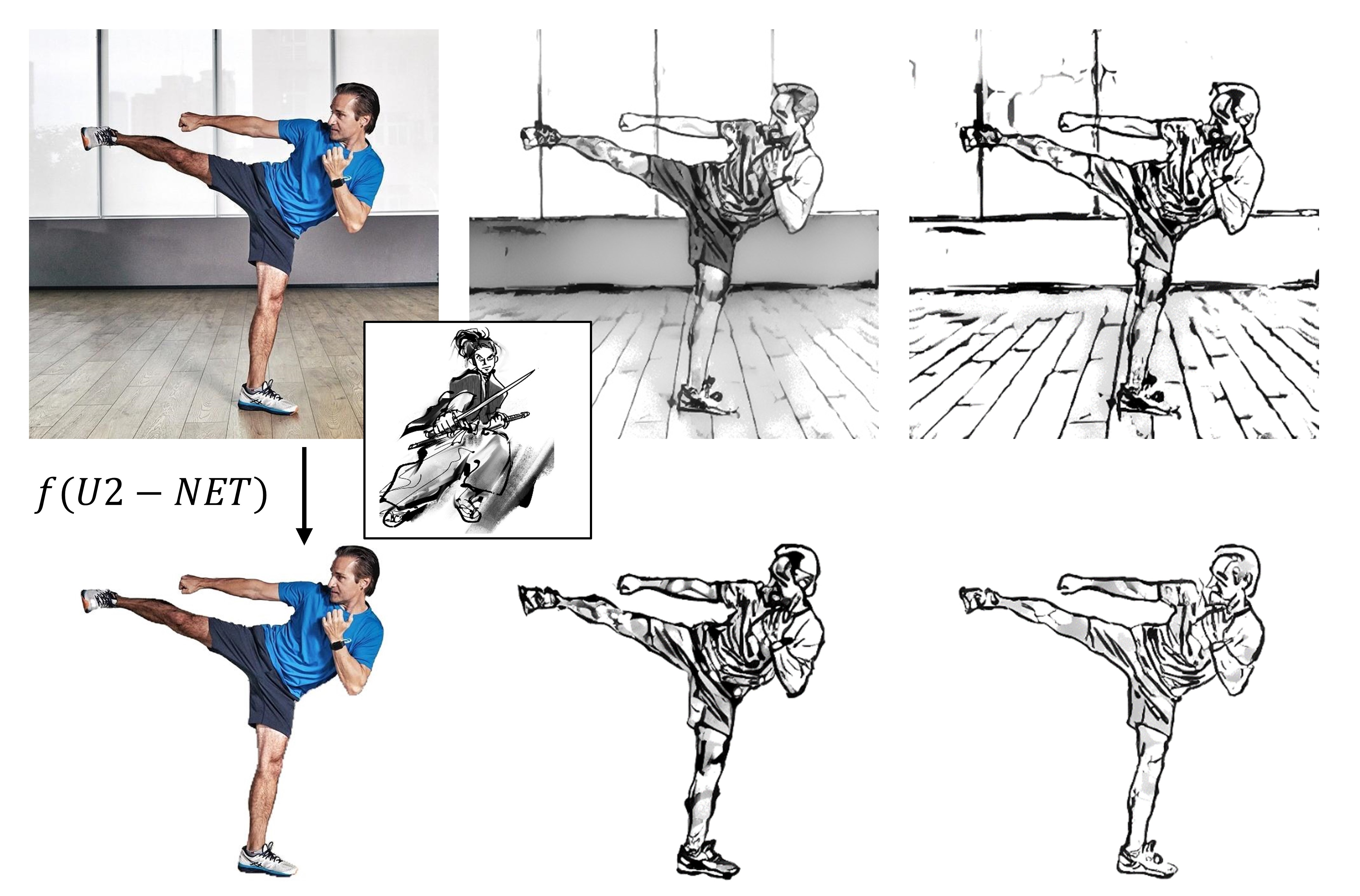}
  \caption{Illustration of sketch extraction with and without foreground extraction using U2-Net~\cite{qin2020u2}. The first row displays sketches that include background details, while the second row presents sketches that focus solely on the foreground object.}
  \label{fig9}
\end{figure}

As demonstrated in Fig.~\ref{fig9}, the first row shows sketch extraction results without foreground extraction, where background details are included in the sketch. In contrast, the second row illustrates the results after applying U2-Net for foreground extraction. Here, the sketches focus solely on the foreground object, providing a cleaner and more relevant depiction. This capability significantly enhances the utility of our model in practical applications, ensuring that the output aligns more closely with the user's artistic vision.

Foreground extraction is particularly beneficial in artistic applications where the emphasis is on the primary subject, such as portrait drawing or object-focused sketches. Previous studies, including CLIPascene~\cite{vinker2023clipascene}, have highlighted the importance of scene decomposition in sketching. Additionally, research on foreground-background separation has shown its effectiveness in enhancing the clarity and impact of visual representations~\cite{chen2017deeplab,xie2020polarmask}. By allowing users to toggle this feature, our model offers greater flexibility and control over the sketching process, catering to diverse artistic needs and preferences.

\section{Experiments}
\subsection{Experimental Settings}
We implement our method using the state-of-the-art text-to-image Stable Diffusion model, utilizing the publicly available version 1.4 checkpoints hosted on Huggingface\footnote{https://huggingface.co/CompVis/stable-diffusion-v-1-4-original/tree/main}. The images are initially converted into noise maps through DDIM deterministic inversion~\cite{song2020denoising}. Notably, the starting noise map begins without a prompt, and the initial sketch outlines \( z^s_0 \) are set unless specified otherwise. Outlines are generated using the default edge detection method, TEED, unless adjustments are made to suit specific data requirements. The default $\alpha$ value for the ``strokes sparse thresholds'' is set to 0.55. During the sampling process, we employ DDIM sampling~\cite{song2020denoising} with 50 denoising steps, and set the classifier-free guidance to 7.5. The parameters \( \zeta \) and \( \beta \) are adjustable based on specific requirements; by default, \( \zeta \) is set to 0.4 and \( \beta \) to 0.5, although adjustments may be necessary for different checkpoints. MixSA does not require training and can activate targeted strategies based on different types of data. For example, in complex scenes, U2-Net is activated to extract the foreground. For different styles of reference images, we adjust \( \zeta \) and \( \beta \) accordingly. We followed the approach used in Semi-ref2Sketch~\cite{semi2sketch}, determining parameters based on the same validation sets. To ensure consistency, we adjusted the parameters using a single test style across all experiments, ensuring uniformity in the results.

\subsection{Test Dataset}

To comprehensively evaluate MixSA, we conducted tests on three diverse datasets:

\textbf{4SKST~\cite{semi2sketch}:} The 4SKST dataset includes 25 color images, each paired with 4 different sketch styles, totaling 100 sketches. Created by professional artists, it is used for research in sketch-based image generation and understanding. We use all color images for testing.

\textbf{FS2K~\cite{fan2022facial}:} The FS2K dataset contains 5,140 pairs of facial images and sketches, covering various ages, genders, and ethnicities. It supports research in facial recognition and sketch synthesis. We randomly select 1,000 pairs for testing.

\textbf{Anime~\cite{taebum2018anime}:} The Anime Sketch Colorization Pair dataset consists of 140,000 pairs of anime-style sketches and their colored versions, used for training models on colorizing line art. It is widely utilized in image-to-image translation research. We randomly select 1,000 pairs for testing.

\subsection{Evaluation Metrics}

In our study, we adopted five established metrics to conduct a quantitative analysis: Peak Signal-to-Noise Ratio (PSNR)~\cite{wang2004image}, Learned Perceptual Image Patch Similarity (LPIPS)~\cite{zhang2018unreasonable}, Fréchet Inception Distance (FID)~\cite{grigorescu2003contour}, Kernel Inception Distance (KID)~\cite{bińkowski2018demystifying}, and Structural Similarity Index (SSIM)~\cite{zhang2018unreasonable}. These metrics respectively evaluate pixel-level accuracy, perceptual similarity based on neural network features, the distributional similarity between sets of images, unbiased FID alternatives for smaller sample sizes, and the structural similarity of images, focusing on luminance, contrast, and structure retention.

While PSNR is adept at measuring pixel-level differences, it often overlooks finer perceptual nuances crucial to human visual assessment~\cite{wang2004image}. Conversely, FID measures the statistical similarity between sets of images but requires a substantial sample size (e.g., 50k images) to produce reliable results~\cite{borji2019pros}. KID provides a reliable alternative to FID for smaller datasets, and SSIM evaluates structural similarities, emphasizing the importance of luminance, contrast, and structure in image quality.

These evaluation metrics are primarily used to assess image quality in natural scenes. Previous studies on reference sketch extraction lacked specialized evaluation metrics. Therefore, we attempted to include as many relevant metrics as possible, choosing the above metrics for comprehensive comparison and evaluation.

\subsection{Comparison with State-of-the-arts}

\begin{table}[t]
\centering
\caption{Quantitative results of comparison with baselines on the 4SKST dataset~\cite{semi2sketch}. The best score is annotated in bold, while the second-best scores are underlined.}
\begin{tabular}{lccc}
  \toprule
\textbf{Methods} &LPIPS$\downarrow$ & PSNR$\uparrow$ & FID$\downarrow$ \\
 \midrule
Ref2sketch~\cite{ref2sketch} &  0.2192&35.02&115.96 \\
Semi-ref~\cite{semi2sketch} & \underline{0.1271}&\underline{35.58}&\textbf{82.18}   \\
Ours & \textbf{0.1124}&\textbf{35.65}&\underline{89.04}\\
 \midrule
SketchKeras~\cite{sketchkeras} & 0.2112&34.99&97.95 \\
IrwGAN~\cite{xie2021unaligned} & 0.1633&34.51&95.28 \\
\cite{park2020swapping}  & 0.2745&35.04&174.12  \\
MUNIT~\cite{huang2018multimodal}& 0.2582&34.23&144.82  \\
Infor-draw~\cite{chan2022learning} & 0.2130&35.05&146.86  \\
  \bottomrule
\end{tabular}
\label{tab1}
\end{table}

\begin{table}[t]
\centering
\caption{Quantitative results of comparison with baselines on the Anime dataset~\cite{taebum2018anime}.}
\begin{tabular}{lccccc}
  \toprule
\textbf{Methods} &LPIPS$\downarrow$ & PSNR$\uparrow$ & FID$\downarrow$ & KID$\downarrow$& SSIM$\uparrow$\\
\midrule
Ref2sketch &0.4655&11.09&220.41&0.1328&0.7350
 \\
Semi-ref &0.4623&13.24&220.61  &0.0836& 0.7848
  \\
StyleID &0.4734&12.72&251.79&\textbf{0.0414}&0.7327
 \\
Ours&\textbf{0.4214}&\textbf{14.64}&\textbf{217.70}&\underline{0.0510}&\textbf{0.7956}
\\
  \bottomrule
\end{tabular}
\label{tab2}
\end{table}

\begin{figure*}[h!]
    \centering
  \includegraphics[width=0.92\linewidth]{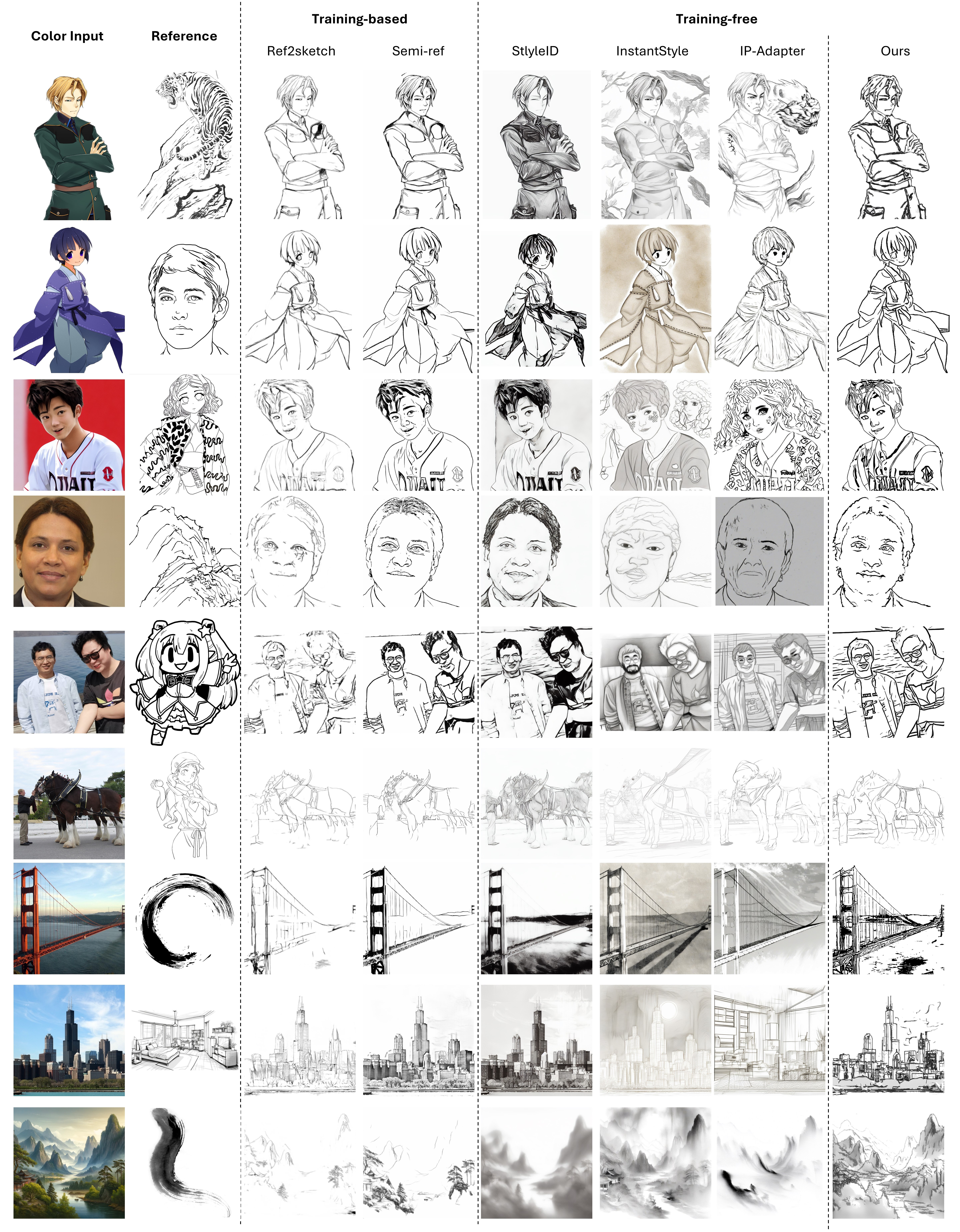}
  \caption{Various examples generated by our method compared to baseline methods. While Semi-ref2sketch produces high-quality results when the input reference sketches are categorized within its training data (as shown in the 2nd and 5th column), they cannot deal with unseen styles that are different from the training data (e.g., 4th, 7th, and 9th rows). Our method generally generates superior results in most cases, effectively replicating the line styles of the input reference sketches.}
  \label{fig10}
\end{figure*}

Here we explore the capabilities of various sketch extraction and style transfer methodologies by benchmarking against five diverse models: Ref2sketch~\cite{ref2sketch}, Semi-ref2sketch~\cite{semi2sketch}, StyleID~\cite{chung2024style}, Instant-Style~\cite{wang2024instantstyle}, and IP-Adapter~\cite{ye2023ip}. Ref2sketch and Semi-ref2sketch primarily focus on extracting sketches from colored images using attention mechanisms and contrastive learning, respectively. These models excel in transforming style attributes by training with specific style guidelines but are limited to the styles available in their training datasets.

On the other hand, StyleID, Instant-Style, and IP-Adapter emerging from the realm of style transfer, offer a training-free and instant adaptation of large-scale diffusion models to preserve artistic styles in text-to-image generation. These models are adept at integrating diverse stylistic elements into generated images, making them highly versatile but less targeted compared to dedicated sketch extraction models.

For comparison, the training parameters and details were determined based on their official code and descriptions in their respective papers. Ref2sketch, being a supervised method, requires a paired dataset, while Semi-ref2sketch, using contrastive learning, requires clustering different sketch styles. We utilized pre-trained weights from their official pages. In the evaluation, methods that can accept a reference image use the same style sketch with an unseen shape image from the 4SKST dataset~\cite{semi2sketch} or Google Images. For InstantStyle and IP-Adapter, we used the official demo of InstantStyle on Huggingface\footnote{https://huggingface.co/spaces/InstantX/InstantStyle}. 
All parameters were set to their official defaults, and the prompt was left empty. We avoided specifying any positive prompt, as terms like ``sketch style'' tended to skew the output away from the reference image’s guidance. Instead, the positive prompt was left empty, and the negative prompt remained at its default settings. Additionally, we used the default ``Load only style blocks'' option for the ``style mode'', as this setting produced results that better aligned with the objectives of the sketch extraction task. 

Fig.~\ref{fig10} shows examples of the generated outputs. We also provide more qualitative comparison in Supplementary. The output sketches produced by these methods were compared to the ground truth using the same three different evaluation metrics employed in the ablation study. The results reported in Table~\ref{tab1} confirm that our method outperforms most of the baselines across these evaluation metrics.

\begin{figure}[t]
  \centering
  \includegraphics[width=\linewidth]{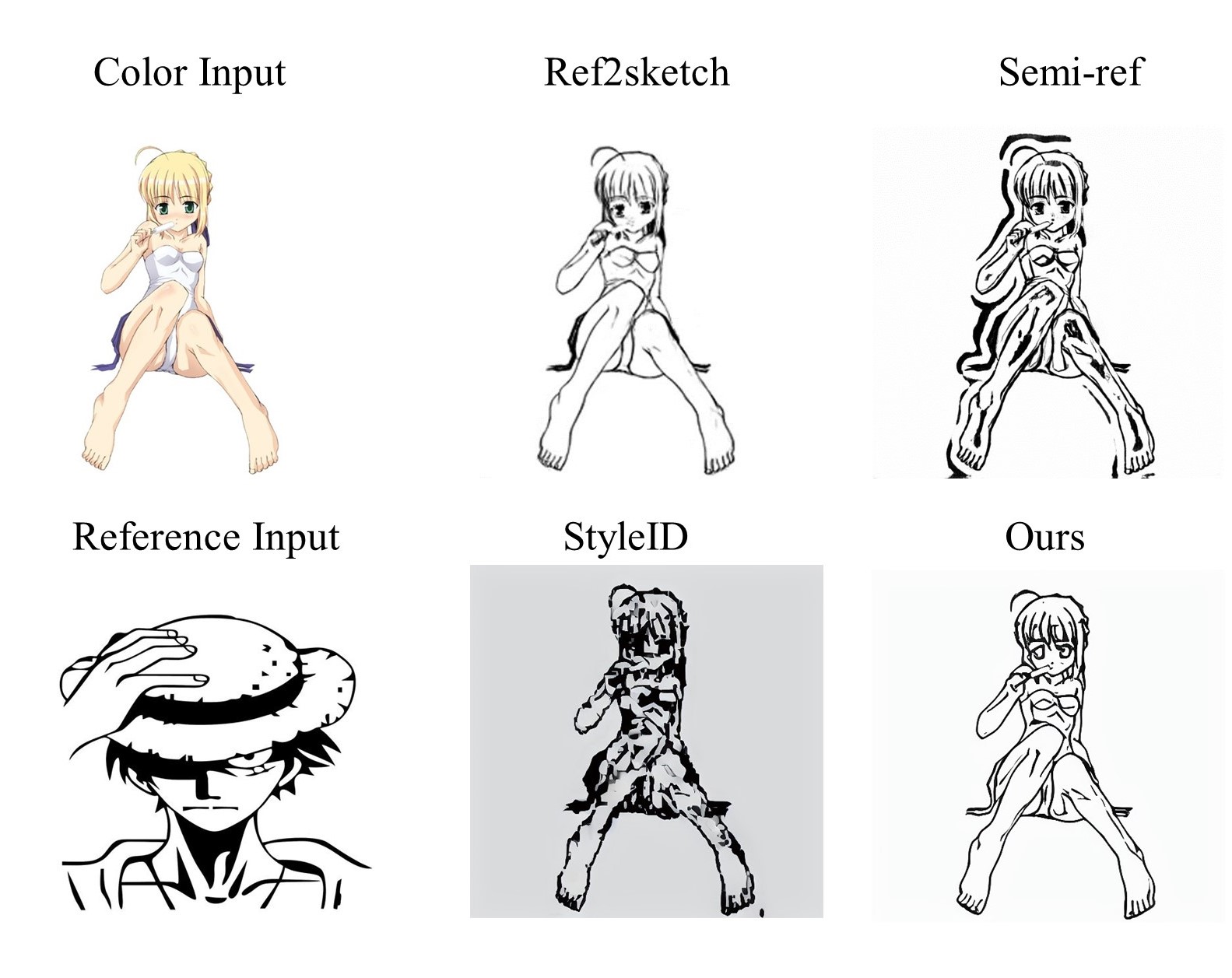}
  \caption{Illustration of sketch extraction on the Anime dataset~\cite{taebum2018anime}. Despite similar training data, Ref2sketch and Semi-ref2sketch models perform poorly in some cases, highlighting the difficulty of clustering sketch styles.}
  \label{fig11}
\end{figure}
\begin{figure}[h]
  \centering
  \includegraphics[width=\linewidth]{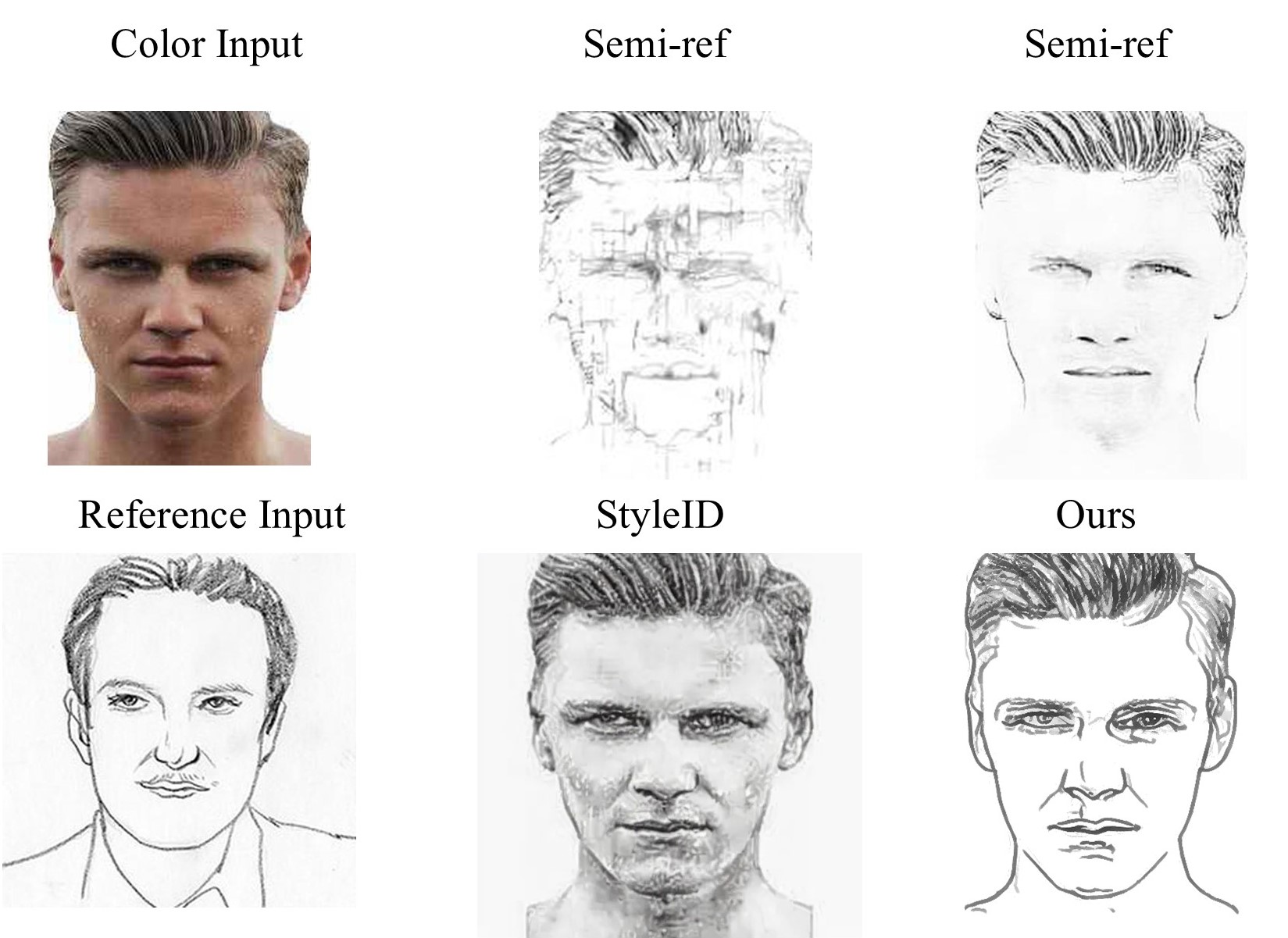}
  \caption{Illustration of sketch extraction on the FS2K dataset~\cite{fan2022facial}. The reference style used here is unseen by the training models (Ref2sketch and Semi-ref2sketch), demonstrating their struggle with generalization.}
  \label{fig12}
\end{figure}

Our model demonstrates superior performance on the Anime dataset (Table~\ref{tab2}) and FS2K dataset (Table~\ref{tab3}) compared to baseline methods. On the Anime dataset, our method achieves the best LPIPS (0.4214), PSNR (14.64), and FID (217.70) scores, along with competitive KID (0.0510) and SSIM (0.7956) scores. Similarly, on the FS2K dataset, our model outperforms others with the best LPIPS (0.4309) and FID (183.37) scores, showcasing its robustness and effectiveness across different datasets.

As shown in Fig.~\ref{fig11}, although the Anime dataset shares similarities with the training data of Ref2sketch and Semi-ref2sketch, these models still perform poorly in some cases. This further validates our initial emphasis on the difficulty of clustering sketch styles. Fig.~\ref{fig12} demonstrates that when the reference style is unfamiliar to the trained models (Ref2sketch and Semi-ref2sketch), they struggle to generalize, despite similar quantitative data in Table~\ref{tab3}. This visual evidence, along with the user study discussed in the next section, highlights the limitations of data-driven training models in handling unseen cases.

\begin{figure*}[h]
  \centering
  \includegraphics[width=\linewidth]{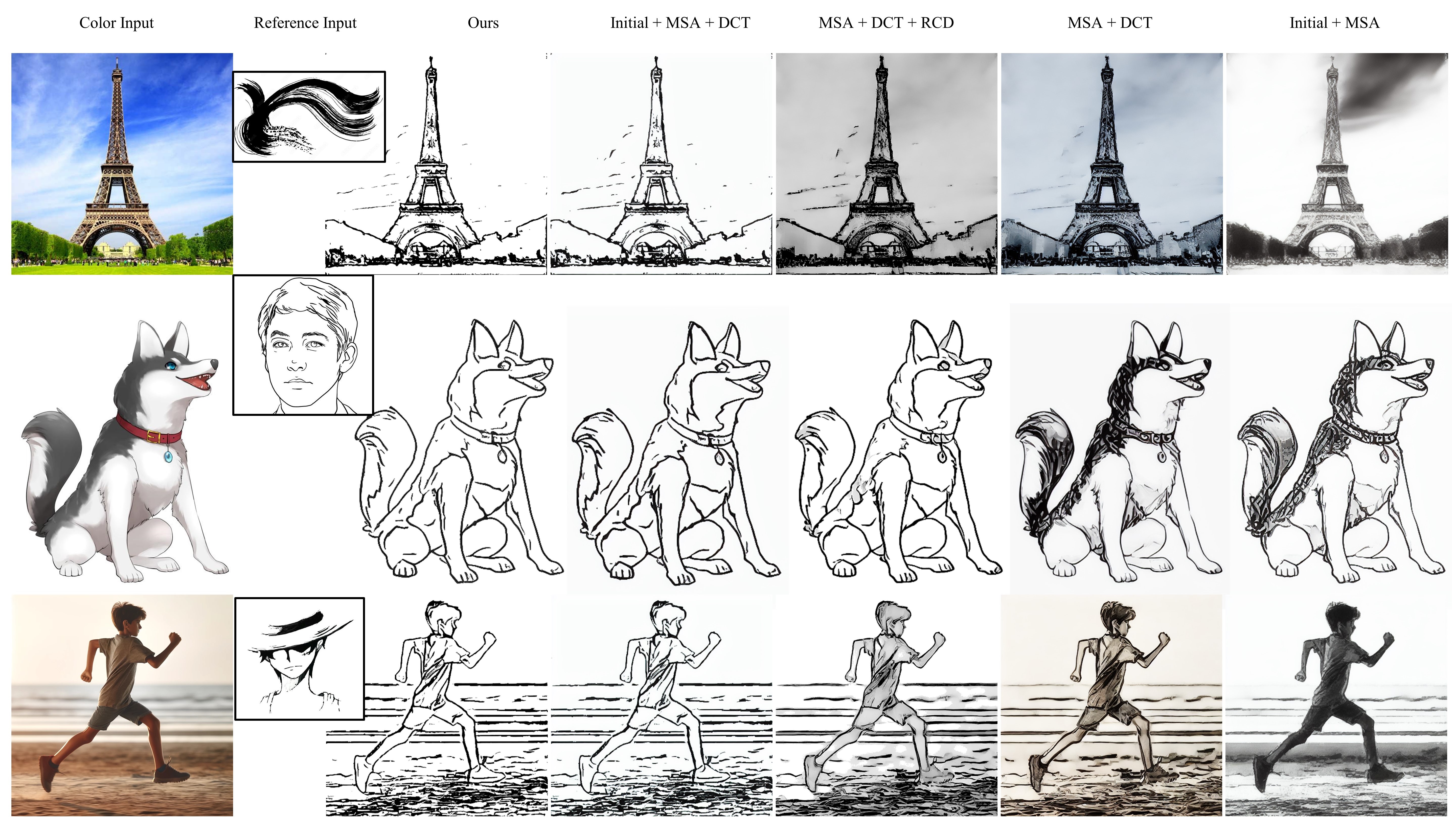}
  \caption{Illustration of sketch extraction in the ablation study. The ``Initial + MSA + DCT + RCD'' configuration achieves the best results, highlighting the importance of each component. Removing the RCD component leads to a noticeable degradation in color fidelity, while excluding the initial contour reduces structural alignment. The MSA component is crucial for style transfer, and the DCT component ensures precise texture and contour handling.}
  \label{fig13}
\end{figure*}

\begin{table}[t]
\centering
\caption{Quantitative results of comparison with baselines and ablation study on FS2K dataset~\cite{fan2022facial}. }
\begin{tabular}{lccc}
  \toprule
\textbf{Methods} &LPIPS$\downarrow$ & PSNR$\uparrow$ & FID$\downarrow$ \\
\midrule
Ours (Initial + MSA + DCT + RCD) & \textbf{0.4309}&28.28&\textbf{183.37} \\
Initial + MSA + DCT & 0.4508&27.98&188.56 \\
MSA + DCT + RCD & 0.4932&26.34&201.40 \\
MSA + DCT & 0.5103&27.94&205.54 \\
Initial + MSA & 0.4720&28.93&197.25\\
\midrule
Ref2sketch &  0.5309&\textbf{29.58}&228.15 \\
Semi-ref & \underline{0.4540}&29.55&185.26   \\
StyleID & 0.5494&28.19&208.64  \\
  \bottomrule
\end{tabular}
\label{tab3}
\end{table}

\subsection{Ablation Study}
To evaluate the contributions of each component in our model, we conducted an ablation study using the FS2K dataset~\cite{fan2022facial}. The components evaluated include the Initial contour, MSA (Mixture Self-Attention), DCT (Decoupling Contour and Texture), and RCD (Reconstruction Color Distribution). The quantitative results are presented in Table~\ref{tab3}.

\textbf{Initial + MSA + DCT + RCD}: This configuration, which includes all components, achieves the best LPIPS (0.4309) scores, indicating the highest perceptual similarity and reconstruction quality. The FID score is also competitive (183.37), showing substantial improvement over most baselines.

\textbf{Initial + MSA + DCT}: Removing RCD from the full model results in a noticeable decline in performance, with LPIPS increasing to 0.4508, PSNR dropping to 27.98, and FID worsening to 188.56. This highlights the importance of the RCD component in maintaining color distribution fidelity during sketch extraction.

\textbf{MSA + DCT + RCD}: Excluding the Initial contour, this configuration shows a moderate LPIPS score (0.4932) and a good PSNR (26.34), but the FID score (201.40) indicates a reduction in overall quality, underscoring the Initial contour’s role in providing structural guidance.

\textbf{MSA + DCT}: Without both the Initial contour and RCD, the model's performance further degrades, with LPIPS at 0.5103, PSNR at 27.94, and FID at 205.54. This demonstrates the significant impact of RCD and the Initial contour in enhancing the model’s capability.

\textbf{Initial + MSA}: When only the DCT and RCD components are removed, the results show a substantial drop in all metrics (LPIPS 0.4720, PSNR 28.93, FID 197.25), indicating that decoupling texture and contour is crucial for maintaining the integrity of sketch details and overall quality.

Comparing our model with existing methods, Ref2sketch and Semi-ref2sketch show competitive results, with Semi-ref2sketch achieving the best FID score (185.26) and a strong LPIPS (0.4540). However, our model with all components (Initial + MSA + DCT + RCD) surpasses these methods in terms of LPIPS and PSNR, demonstrating superior perceptual quality and detail preservation.

In summary, the ablation study confirms that each component of our model contributes significantly to its performance. The Initial contour provides structural alignment, MSA ensures effective style transfer, DCT allows for precise texture and contour handling, and RCD addresses color distribution challenges inherent in sketch generation (Fig.~\ref{fig13}).

\begin{table}[t]
\centering
\caption{Results from the user perceptual study.}
\begin{tabular}{lc}
  \toprule
\textbf{Methods} & User Score\\
\midrule
Ref2sketch~\cite{ref2sketch}  &  9.60\% \\
Semi-ref2sketch~\cite{semi2sketch} & 13.37\%  \\
StyleID~\cite{chung2024style} & 5.35\% \\
Ours & \textbf{71.68\%} \\
  \bottomrule
\end{tabular}
\label{tab4}
\end{table}

\subsection{User Study}
We conducted a user study with 50 participants to evaluate preferences regarding the strength of reference sketch guidance and the overall artistic visual effects produced by our model. In this survey, we compared the outputs of our model with those of three other state-of-the-art models. Each model generated sketches from 100 colored images based on three reference sketches. For each participant, 20 randomly selected images were presented, and they were asked to choose the best sketch from the four models' outputs that most accurately represented the reference sketch's brush strokes and overall effect.

The results of the study, summarized in Table~\ref{tab4}, indicate a clear preference for our model's outputs. Participants overwhelmingly favored our model, with 71.68\% selecting it as producing the most faithful and visually appealing sketches compared to the reference sketch style. This high preference score underscores the effectiveness of our model in adhering to the stylistic nuances of the reference sketches, offering more accurate and artistically satisfying results.

\begin{figure}[t]
  \centering
  \includegraphics[width=\linewidth]{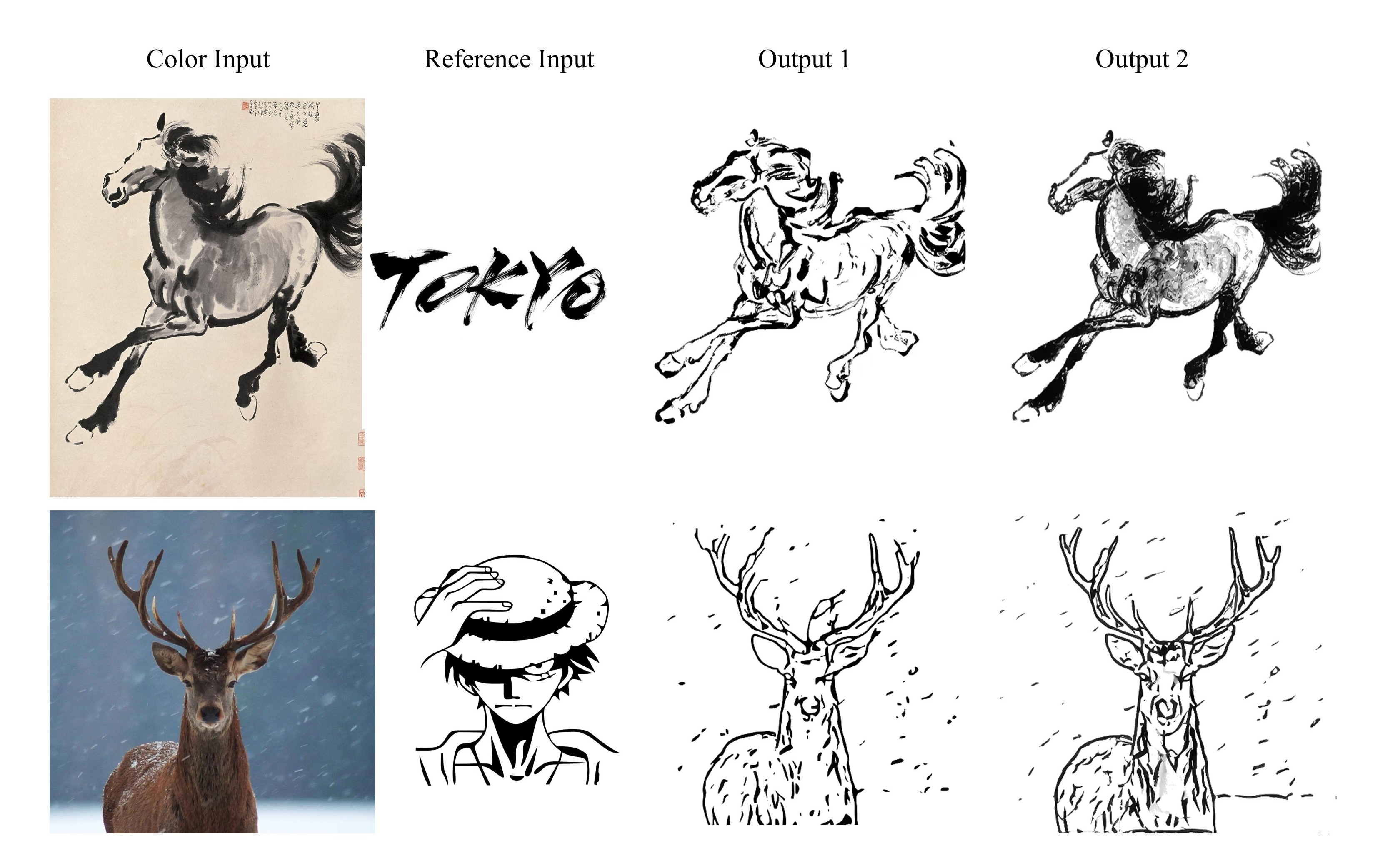}
  \caption{Art drawing generated by adjusting the abstraction levels of sketch and texture: (Output 1: $\zeta = 0.5$, $\beta = 0.04$; Output 2: $\zeta = 0.5$, $\beta = 0.5$).}
  \label{fig14}
\end{figure}
\begin{figure}[t]
  \centering
  \includegraphics[width=\linewidth]{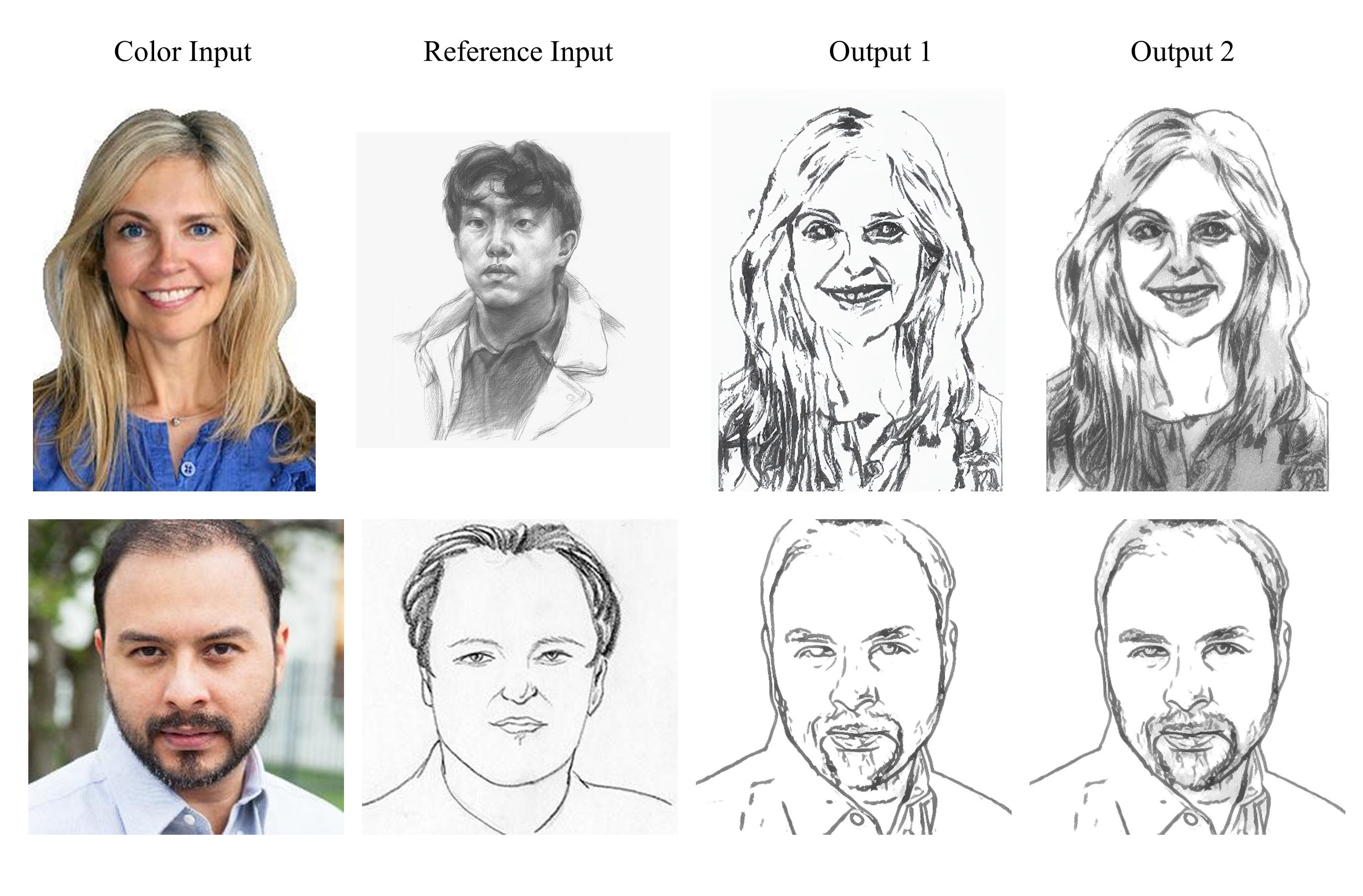}
  \caption{Portrait drawing generated by adjusting the abstraction levels of sketch and texture: (Output 1: $\zeta = 0.5$, $\beta = 0.04$; Output 2: $\zeta = 0.5$, $\beta = 0.5$).}
  \label{fig15}
\end{figure}

\section{More Applications}

\subsection{Art Drawing}

The application of automated sketch extraction in artistic creation holds tremendous potential. Our model leverages reference sketches and allows for the adjustment of abstraction levels, texture retention, and stylistic elements, providing artists with unprecedented flexibility and control over their creative process. This capability significantly reduces the manual effort traditionally required to add textures or adjust strokes, empowering artists to produce new works of art that seamlessly integrate various styles.

Examples are shown in Fig.~\ref{fig14}. By manipulating parameters such as \(\zeta\) and \(\beta\), users can dynamically fine-tune the output to achieve a spectrum of artistic effects, ranging from highly abstract representations to detailed and textured sketches. This versatility not only facilitates the creation of unique artistic expressions but also enables the generation of consistent datasets of sketches in the same style, which can be invaluable for further training and refinement of generative models or for use in stylistic studies.

Moreover, our model’s ability to emulate different brushwork styles, inspired by traditional techniques~\cite{cahill1959tao} like \textit{Xieyi} (freehand brushwork) and \textit{Gongbi} (fine brushwork), opens new avenues for preserving and innovating within these artistic traditions. Artists can experiment with blending these techniques or explore entirely new styles, all while maintaining control over the finer details of their work.

The integration of DDIM inversion techniques ensures that the essential features of the reference sketches are preserved and accurately transferred, enhancing the model’s reliability and effectiveness. This approach not only captures the visual aesthetics of the reference images but also adheres to the structural nuances that define different artistic styles.

\subsection{Portrait Drawing}

Our model demonstrates significant potential for applications in portrait drawing, as illustrated in Fig.~\ref{fig12} and Fig.~\ref{fig15}. The flexibility to adjust texture and brush stroke styles addresses a critical demand in portrait art, where capturing the unique features and expressions of subjects requires a high degree of customization. Traditional models, trained on fixed datasets, typically generate a limited range of outputs, constraining artists to predefined styles. In contrast, our model enables users to dynamically select from a diverse array of stylistic interpretations, providing the ability to fine-tune the nuances of texture and stroke to match personal or client preferences.

This capability is particularly beneficial in portrait drawing, where the emphasis on individual detail and stylistic variation is paramount. Artists and users can experiment with different levels of abstraction and realism, from detailed, fine brushwork to more expressive, freehand styles. By adjusting parameters such as \(\zeta\) and \(\beta\), users can achieve a balance between maintaining the integrity of facial features and introducing artistic flair, ensuring each portrait is both recognizable and artistically compelling.

Furthermore, the ability to produce multiple stylistic variations from a single input allows for broader creative exploration, fostering innovation in portrait art. This flexibility not only enhances the artistic process but also improves user satisfaction by offering a wide range of artistic possibilities, making the model an invaluable tool for both professional artists and enthusiasts alike.

\begin{figure}[t]
  \centering
  \includegraphics[width=\linewidth]{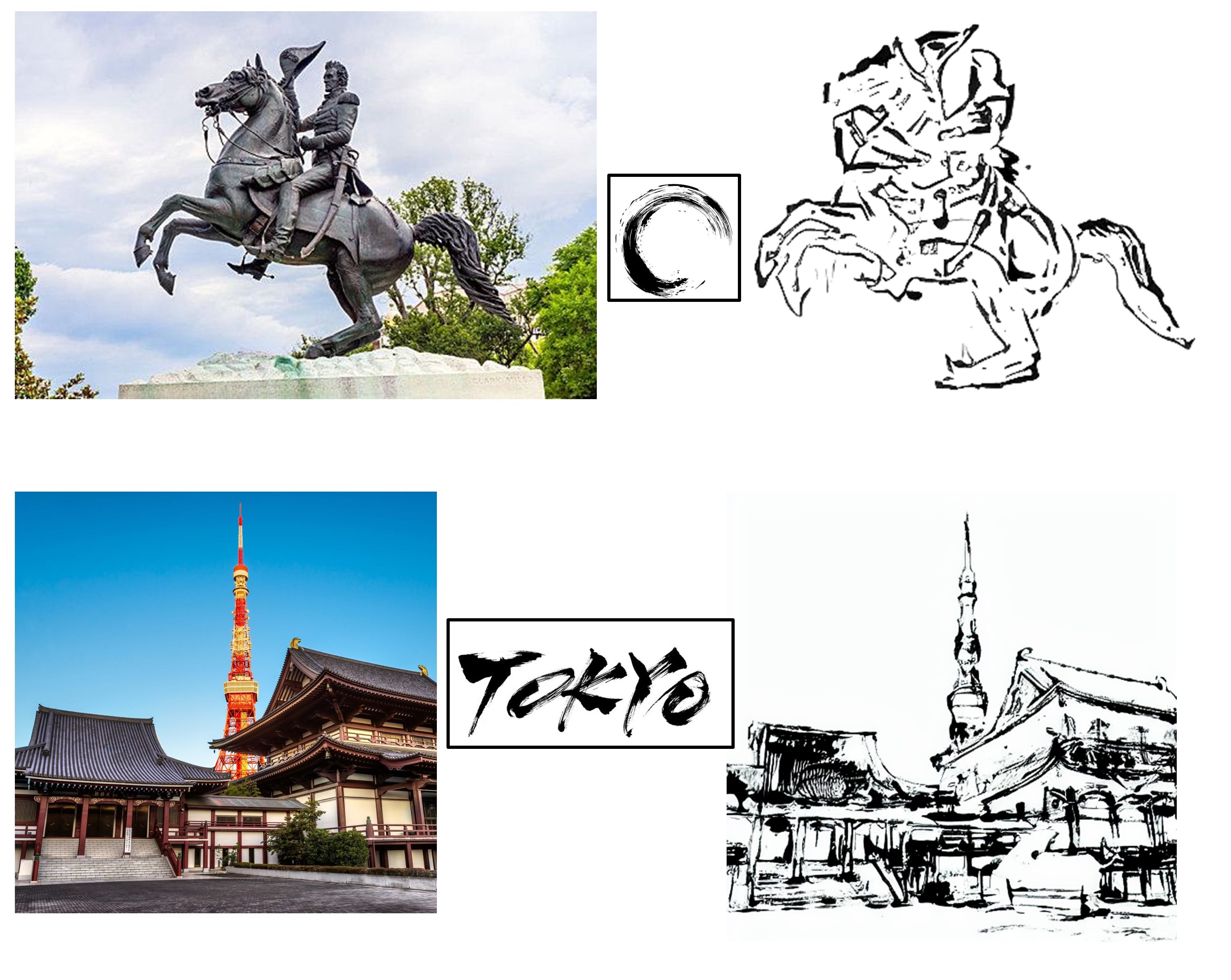}
  \caption{Examples of failure cases where light or low-contrast strokes in the reference sketch led to poor reconstruction of textures and brush stroke details.}
  \label{fig16}
\end{figure}

\section{Conclusion and Discussion}

Our paper introduces a novel sketch extraction method without the need for additional training. This method allows the model to learn and replicate the brush stroke style of a reference image for sketch extraction. By fully utilizing the features of the attention layers, we artistically re-render a given colored image according to the stroke style and aesthetic of the reference image. Additionally, by observing the diverse representation of textures in different sketches, we propose a flexible strategy for texture placement, decoupling the texture and contours of the object. Through these innovations, our method captures a wide range of brush stroke styles, producing results that satisfactorily emphasize styled sketch extraction from colored images, reflecting the varied artistic interpretations inherent in different styles.

\textbf{Limitations and Future Work.} Despite the promising results achieved using diffusion models for sketch extraction, several limitations persist, primarily due to the challenges in reconstructing color textures and details with a limited number of sampling steps. This issue is particularly evident when dealing with reference sketches that have faint or low-contrast strokes. For instance, as illustrated in Fig.~\ref{fig16}, the sketch extraction for a sculpture resulted in reduced overall recognizability. After using the pre-trained U2-Net~\cite{qin2020u2} to separate the foreground, the generated sketch applied overly heavy strokes to the horse's head and the human face, obscuring finer details. Similarly, when attempting to render a scene with brush strokes emulating traditional Chinese painting styles, the expectation was for elements like the tower to be captured with a single fluid stroke. However, the output sketch was significantly constrained by the initial contour information, failing to achieve the desired artistic effect.

These examples highlight the need for further refinement in rendering brush stroke styles. Future research could focus on optimizing stroke style rendering to better match the intricacies of reference sketches. Improving the model’s ability to adaptively control stroke weight and texture detail could enhance the fidelity of the extracted sketches.
Additionally, integrating more sophisticated post-processing techniques and using additional training data to fine-tune the model's response to various artistic styles could mitigate these limitations. Enhancing the diffusion model's capability to preserve subtle details through better noise handling and more effective inversion processes will also be crucial.

\bibliographystyle{IEEEtran}
\bibliography{ref}

\begin{IEEEbiography}[{\includegraphics[width=1in,height=1.25in,clip,keepaspectratio]{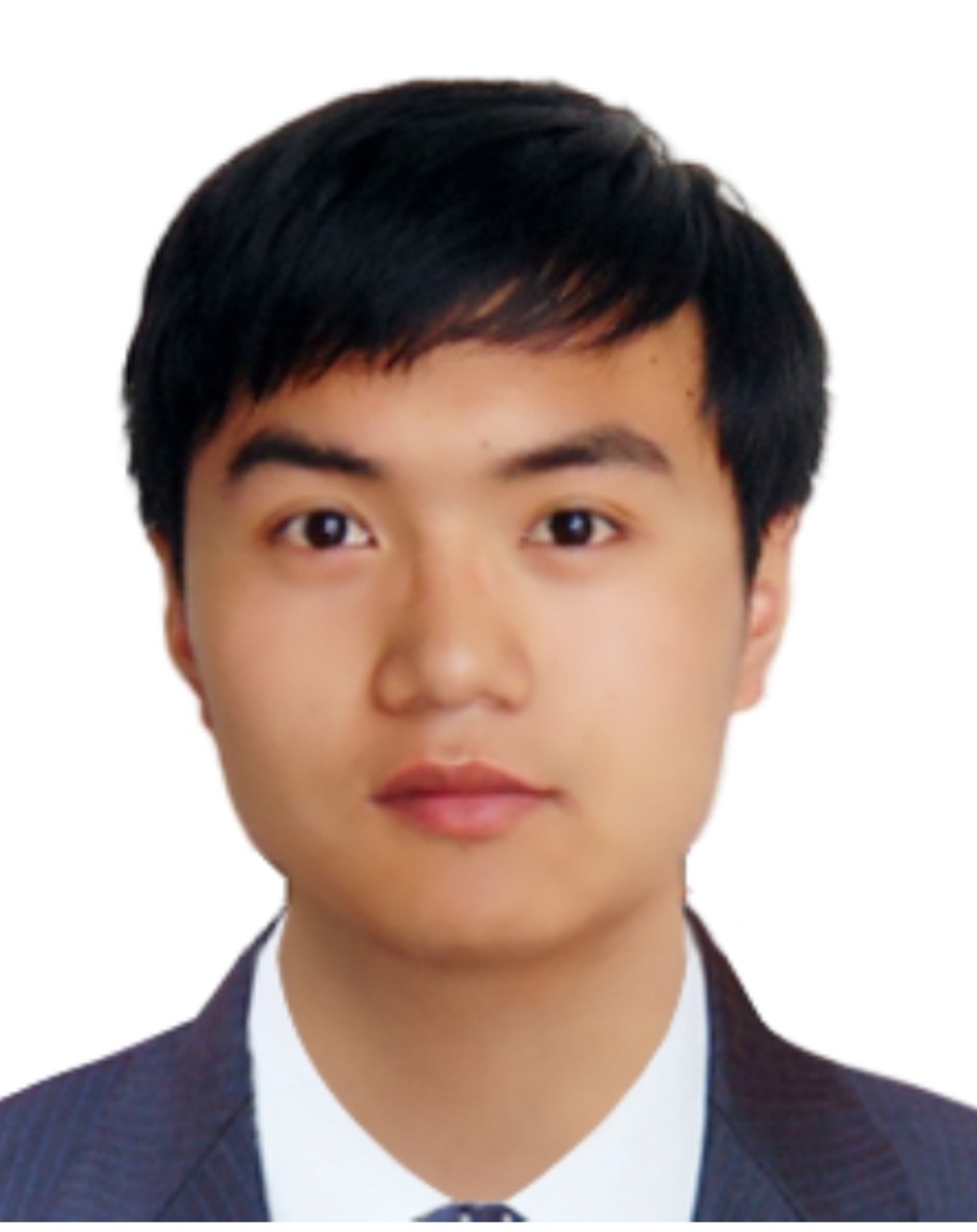}}]{Rui Yang (Student Member, IEEE)}
received an M.S.Eng. degree in Computer Science and Engineering from Xi'an Jiaotong University, Xi'an, China, in 2019. He is currently pursuing a Ph.D. degree in Computer Science at Shaanxi Normal University, Xi’an, China. His research interests include semantic segmentation, image generation, image processing, computer vision, and machine learning. 
\end{IEEEbiography}
\begin{IEEEbiography}[{\includegraphics[width=1in,height=1.25in,clip,keepaspectratio]{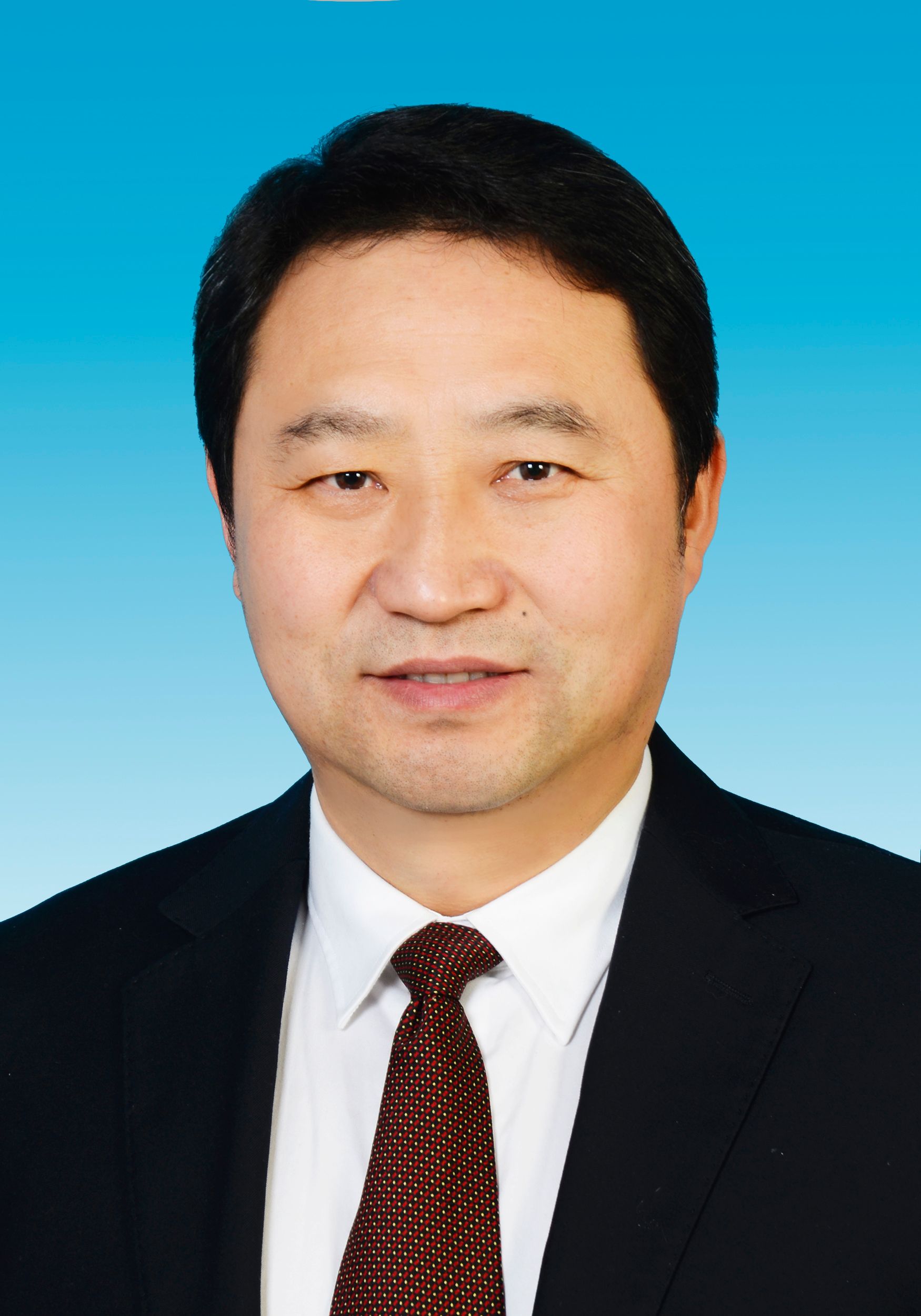}}]{Xiaojun Wu}
received a B.S. in Computer Science and Engineering from Xi'an Jiaotong University in 1993 and obtained a Ph.D. degree in system engineering from Northwestern Polytechnical University, Xi’an, China, in 2005. He served as a visiting scholar at the University of Illinois Chicago in the United States. Currently, he is a professor and doctoral supervisor at Shaanxi Normal University. He has previously worked at institutions such as Huawei Technologies Co., Ltd., and Northwestern Polytechnical University, engaging in technical research and teaching. His research interests encompass artificial intelligence, big data, sensor networks, and the theory and applications of complex systems.
\end{IEEEbiography}
\begin{IEEEbiography}[{\includegraphics[width=1in,height=1.25in, clip,keepaspectratio]{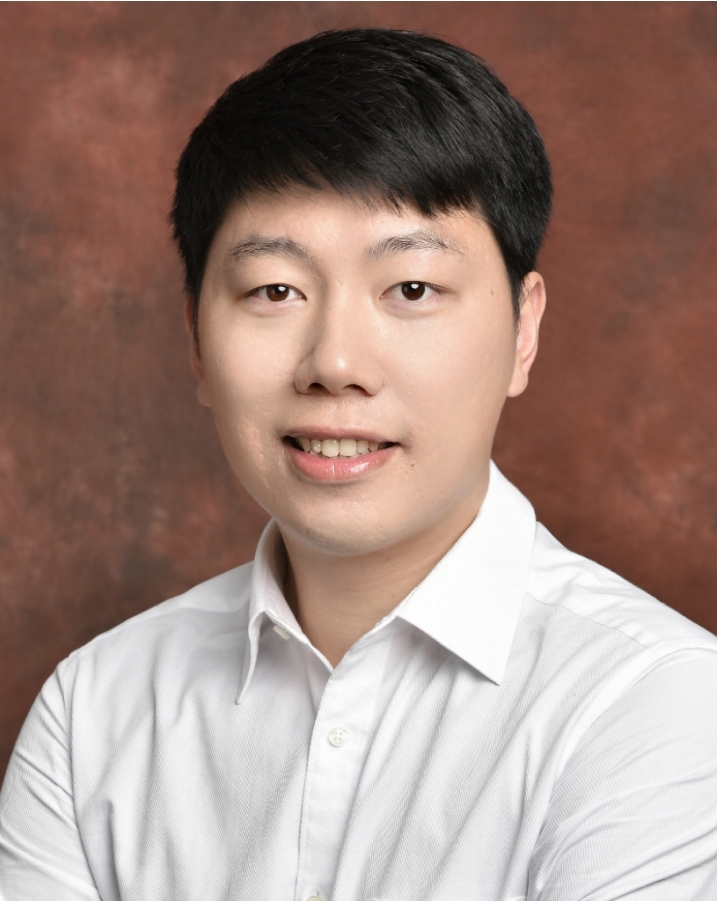}}]{Shengfeng He (Senior Member, IEEE)} is an associate professor in the School of Computing and Information Systems, Singapore Management University. He was on the faculty of the South China University of Technology, from 2016 to 2022. He obtained B.Sc. and M.Sc. degrees from Macau University of Science and Technology in 2009 and 2011 respectively, and a Ph.D. degree from City University of Hong Kong in 2015. His research interests include computer vision and generative models. He has won awards such as Google Research Awards, the Best Paper Award in PerCom24, and Lee Kong Chian Fellowship. He is a senior member of IEEE and CCF. He serves as the lead guest editor of the IJCV, the associate editor of IEEE TNNLS, IEEE TCSVT, Visual Intelligence, and Neurocomputing. He also serves as the area chair/senior program committee of ICML, AAAI, IJCAI, and BMVC.
\end{IEEEbiography}
\vfill
\include{supplementary_material}

\end{document}

%% file: supplementary_material.tex
\pagenumbering{roman}
\setcounter{figure}{0}
\setcounter{section}{0}
\renewcommand{\thefigure}{A\arabic{figure}}
\renewcommand{\thesection}{A\roman{section}}

\clearpage
\twocolumn[
    \begin{center}
        \Large \textbf{MixSA: Training-free Reference-based Sketch Extraction via Mixture-of-Self-Attention\\ --Supplementary Materials--}
    \end{center}
]

\section{Analysis of Trade-off Parameters}\label{a1}

\begin{figure*}[t]
    \centering
    \includegraphics[width=0.95\textwidth]{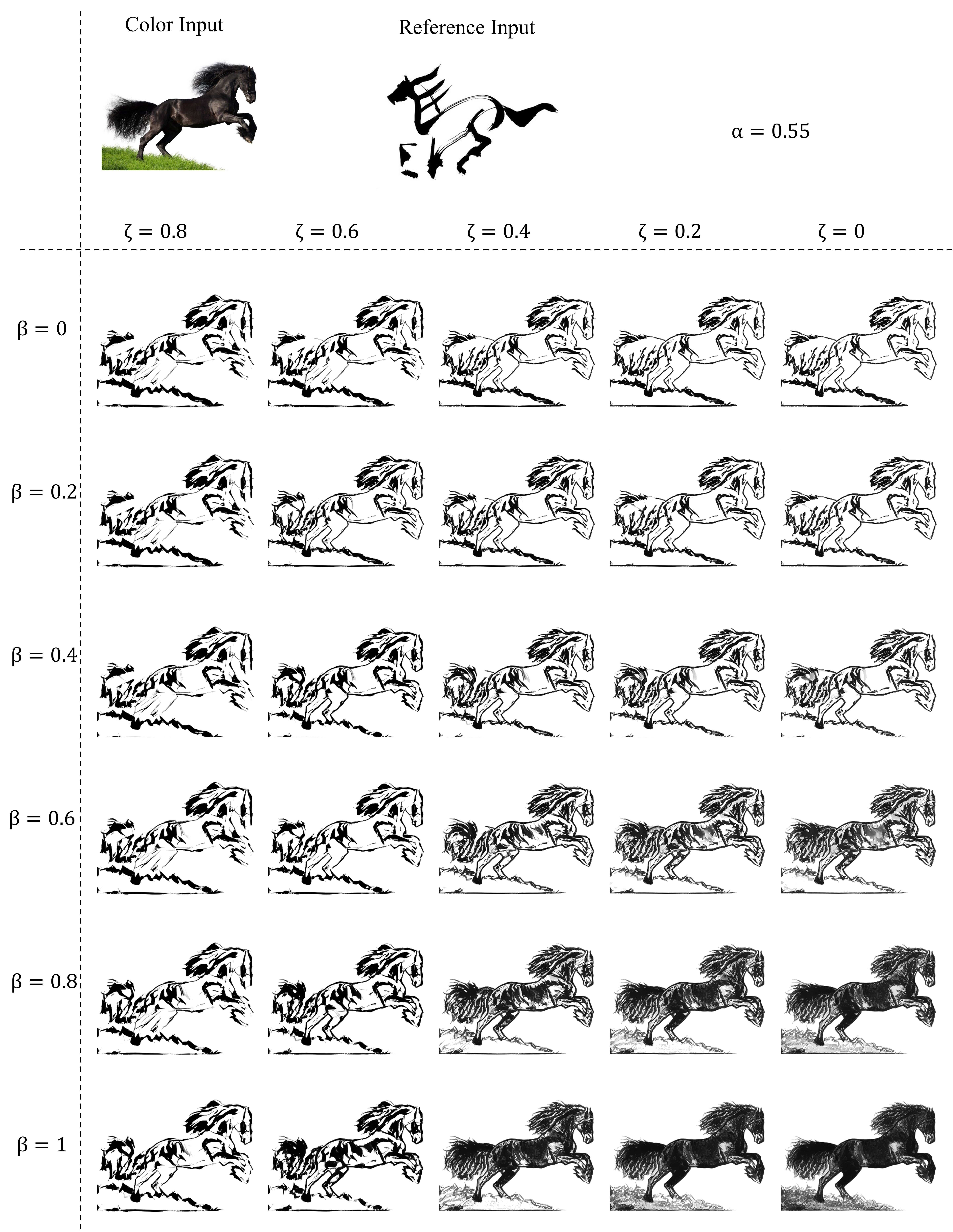}
    \caption{Effects of varying parameters $\zeta$ (reference alignment) and $\beta$ (texture retention) on the output sketches. Increasing $\zeta$ results in higher adherence to the reference sketch, while increasing $\beta$ retains more texture from the color input. Zoom in to view details.}
    \label{A2}
\end{figure*}

\begin{figure*}[t]
    \centering
    \includegraphics[width=0.95\textwidth]{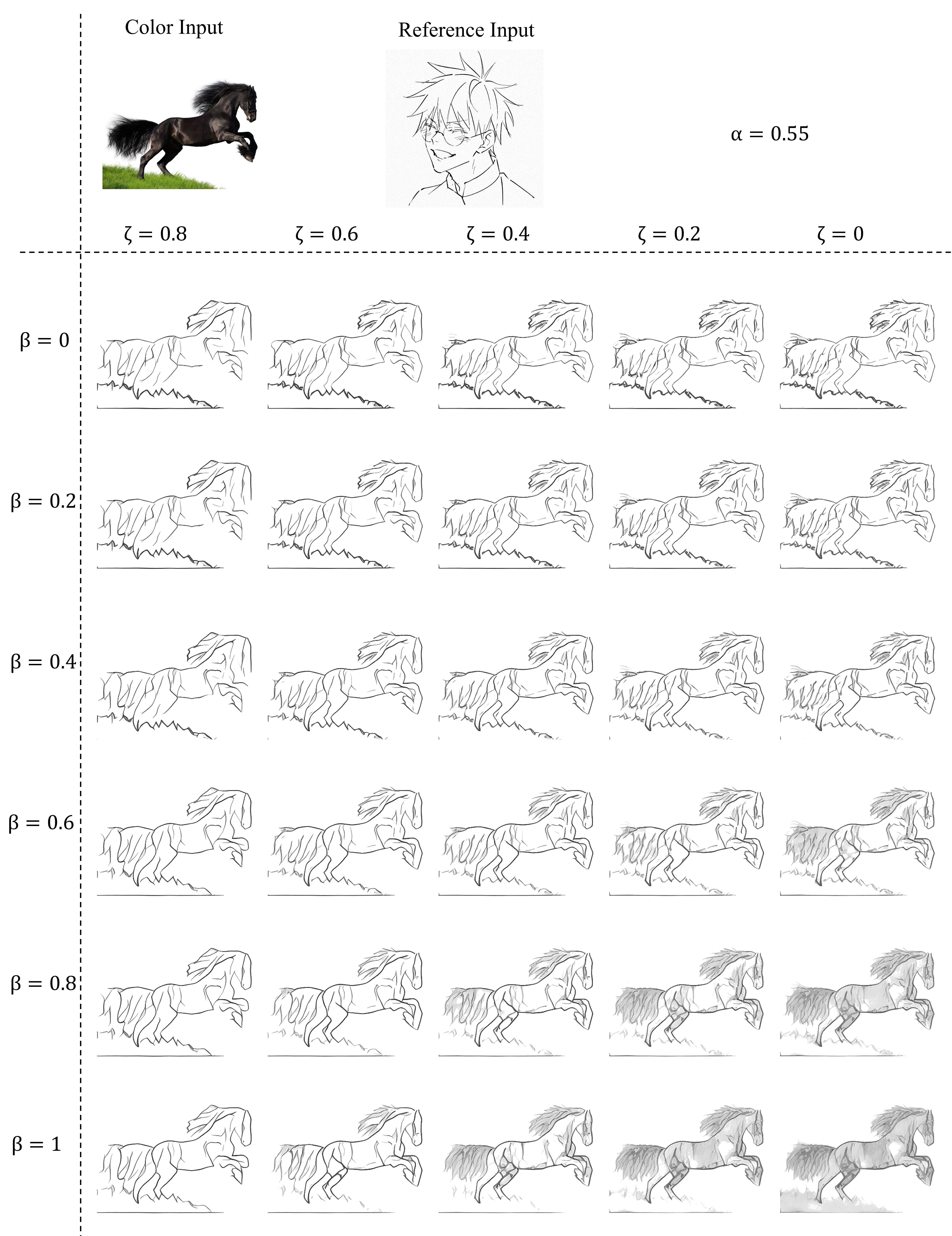}
\caption{Impact of varying $\zeta$ (reference alignment) and $\beta$ (texture retention) on sketch output. Higher $\zeta$ increases reference sketch influence, while higher $\beta$ retains more texture from the original color input. Zoom for details.}
    \label{A2_1}
\end{figure*}

By adjusting the parameters $\zeta$ and $\beta$, sketches can be generated with varying degrees of texture retention, reference adherence, and sparsity. The parameter $\zeta$ controls the alignment with the reference sketch, while $\beta$ modulates the level of texture from the original image. Figs.~\ref{A2} and~\ref{A2_1} demonstrate the effects of different $\zeta$ and $\beta$ combinations on the output sketches.




\begin{figure}[h]
    \centering
    \includegraphics[width=0.5\textwidth]{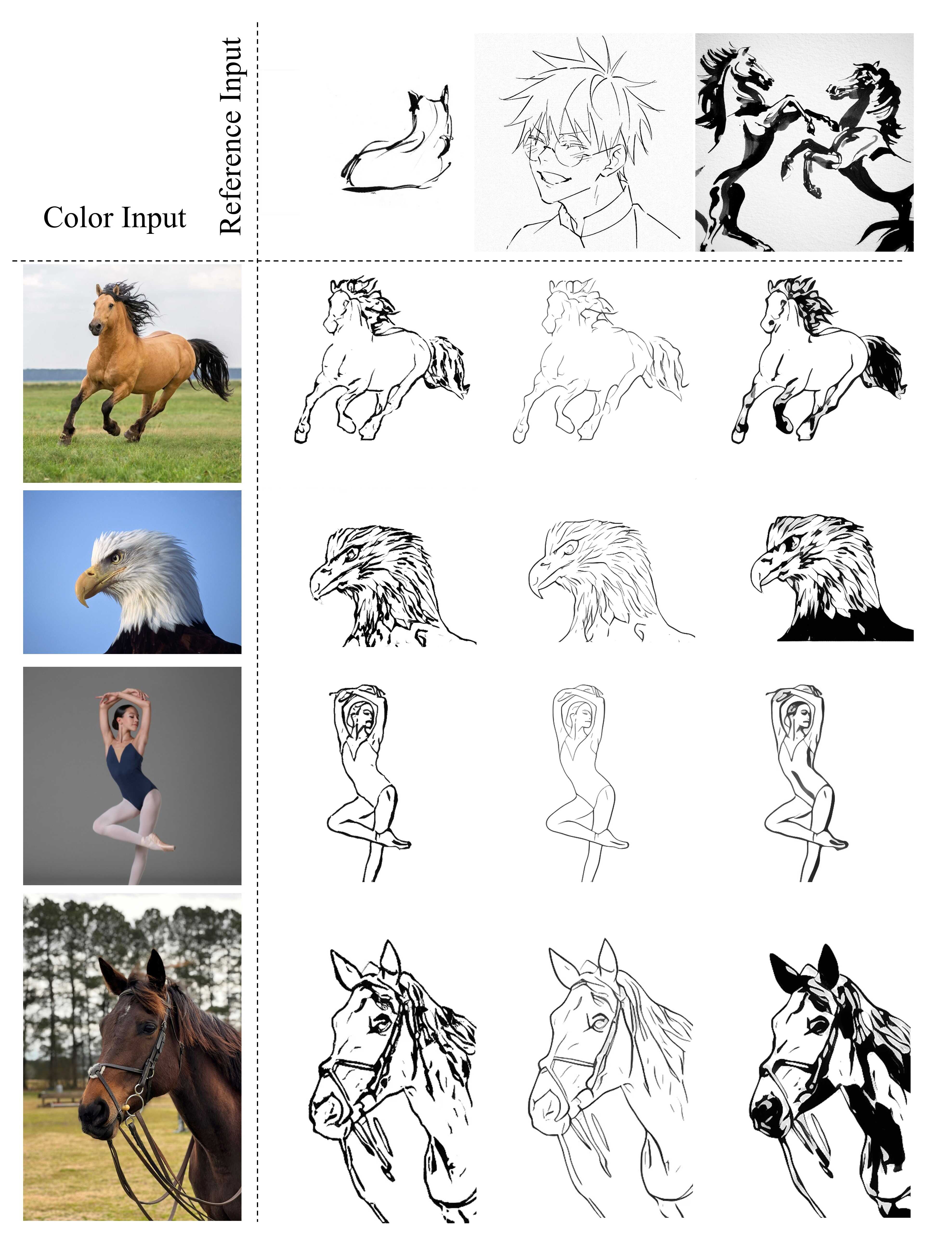}
    \caption{Examples of the reference sketch styles used in our user study, demonstrating a variety of brush strokes and artistic effects across different color inputs.}
    \label{A1}
\end{figure}

\section{User Study Details}\label{a3}

We conducted a user study with 50 participants to evaluate preferences regarding the strength of reference sketch guidance and the overall artistic effects produced by our model. Participants were presented with sketches generated by our method and three other state-of-the-art models, each based on 100 color images and different reference sketches (see Fig.~\ref{A1}). Visual comparisons of the outputs from each model are provided in Figs.~\ref{A3}--\ref{A8}, demonstrating the effectiveness of our approach. 

\section{Additional Qualitative Analysis}\label{a4}

\begin{figure*}[t]
  \centering
  \includegraphics[width=0.9\linewidth]{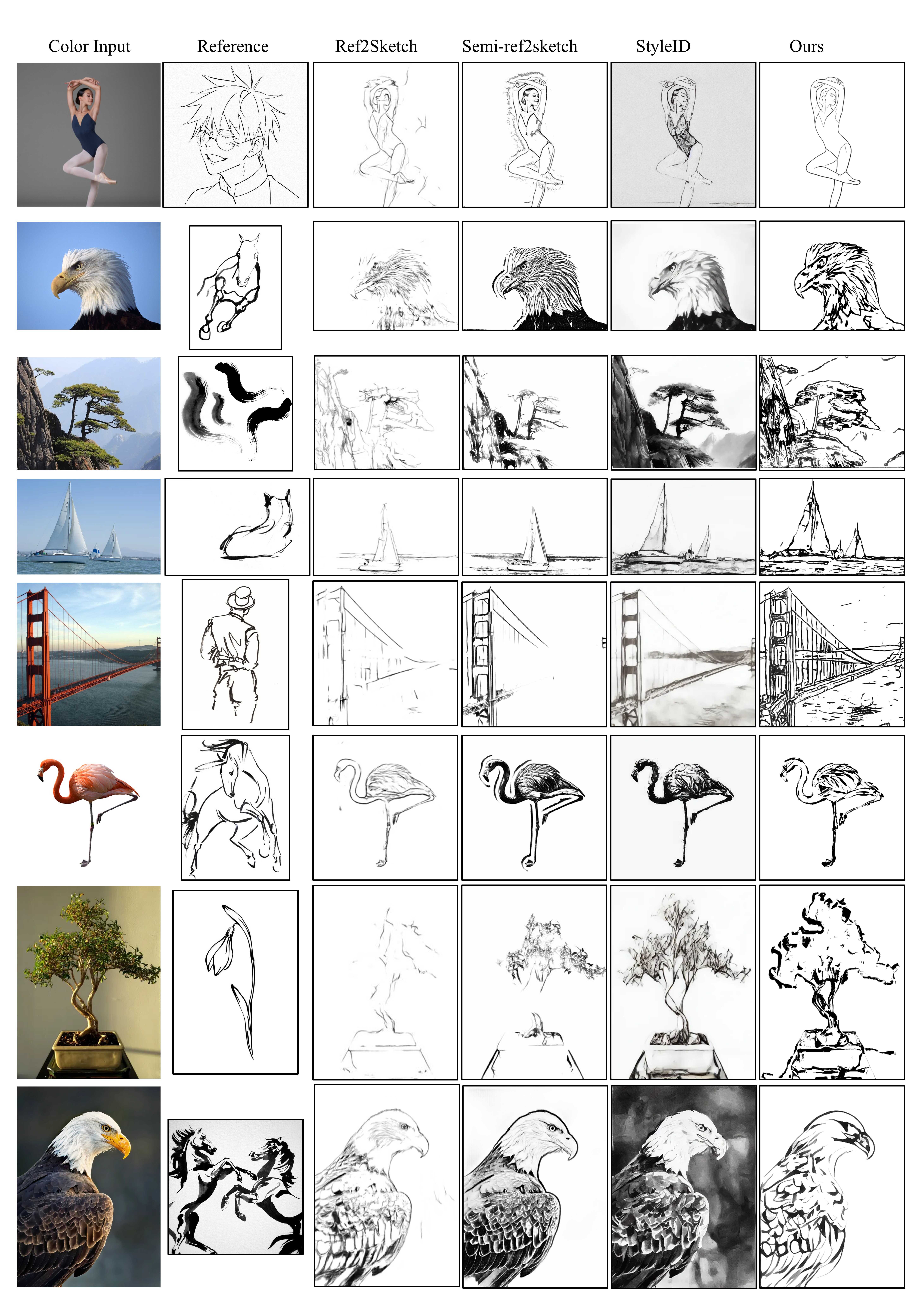}
  \caption{Comparison of different models with various reference sketches. Our model consistently surpasses state-of-the-art methods (Ref2Sketch, Semi-ref2sketch, and StyleID) in preserving the reference sketch's brush strokes and artistic style across various input images.}
  \label{A3}
\end{figure*}

\begin{figure*}[t]
  \centering
  \includegraphics[width=0.9\linewidth]{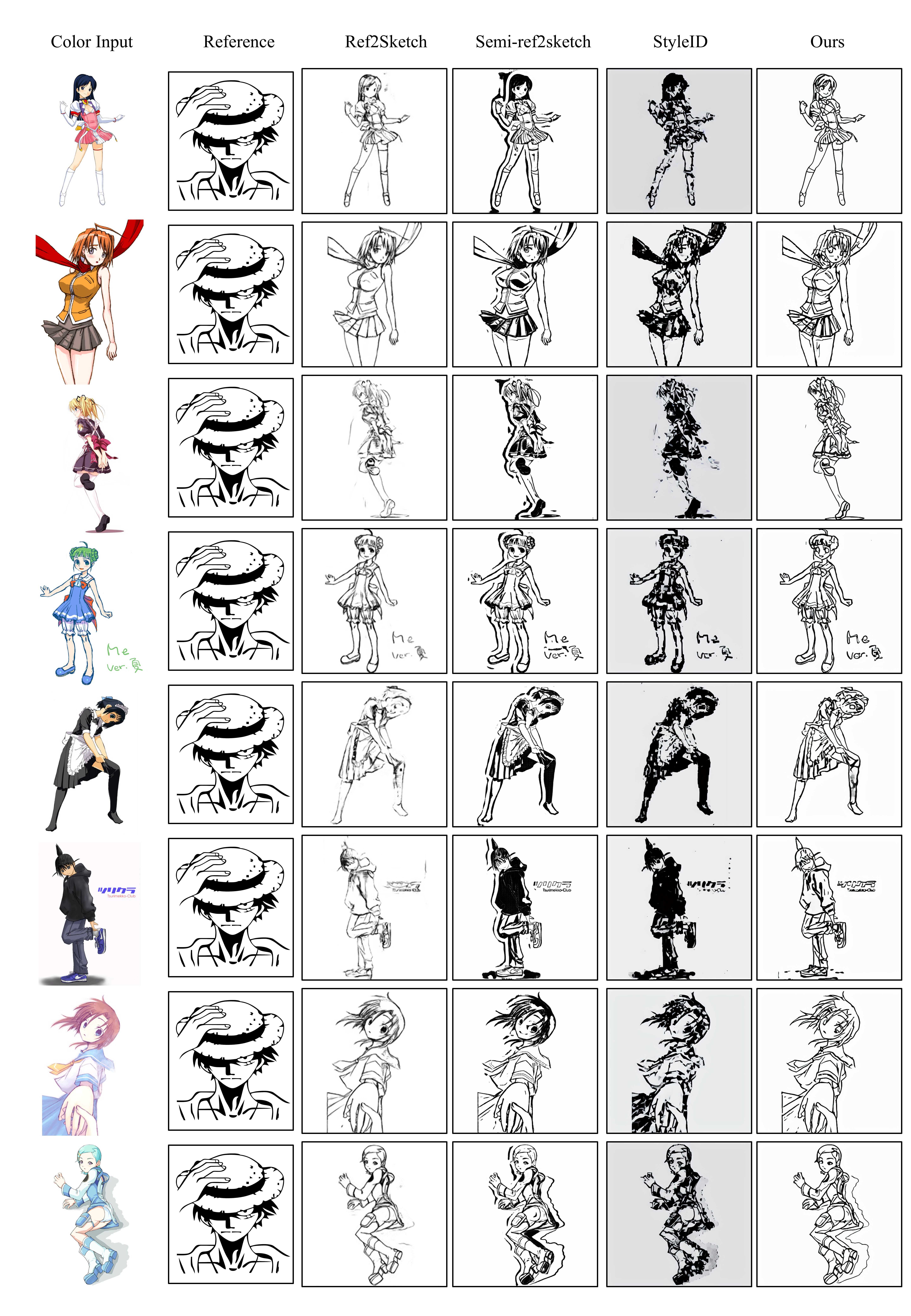}
  \caption{Further comparison results with baselines (Ref2Sketch, Semi-ref2sketch, and StyleID) using the same reference sketch. Zoom in for viewing details.}
  \label{A6}
\end{figure*}

\begin{figure*}[t]
  \centering
  \includegraphics[width=0.9\linewidth]{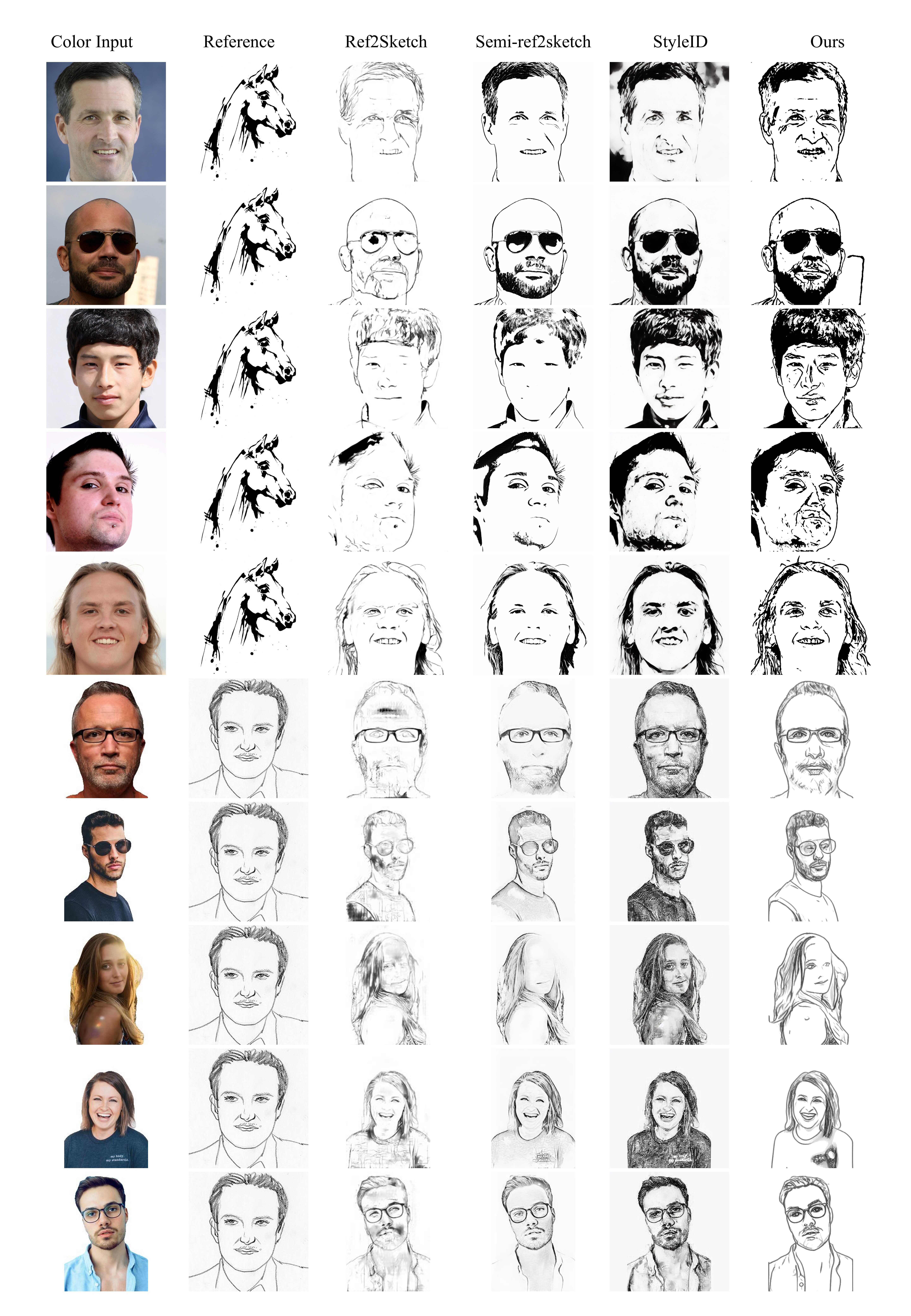}
  \caption{Additional comparison with the same reference sketch. Zoom in for viewing details.}
  \label{A7}
\end{figure*}

\begin{figure*}[t]
  \centering
  \includegraphics[width=0.9\linewidth]{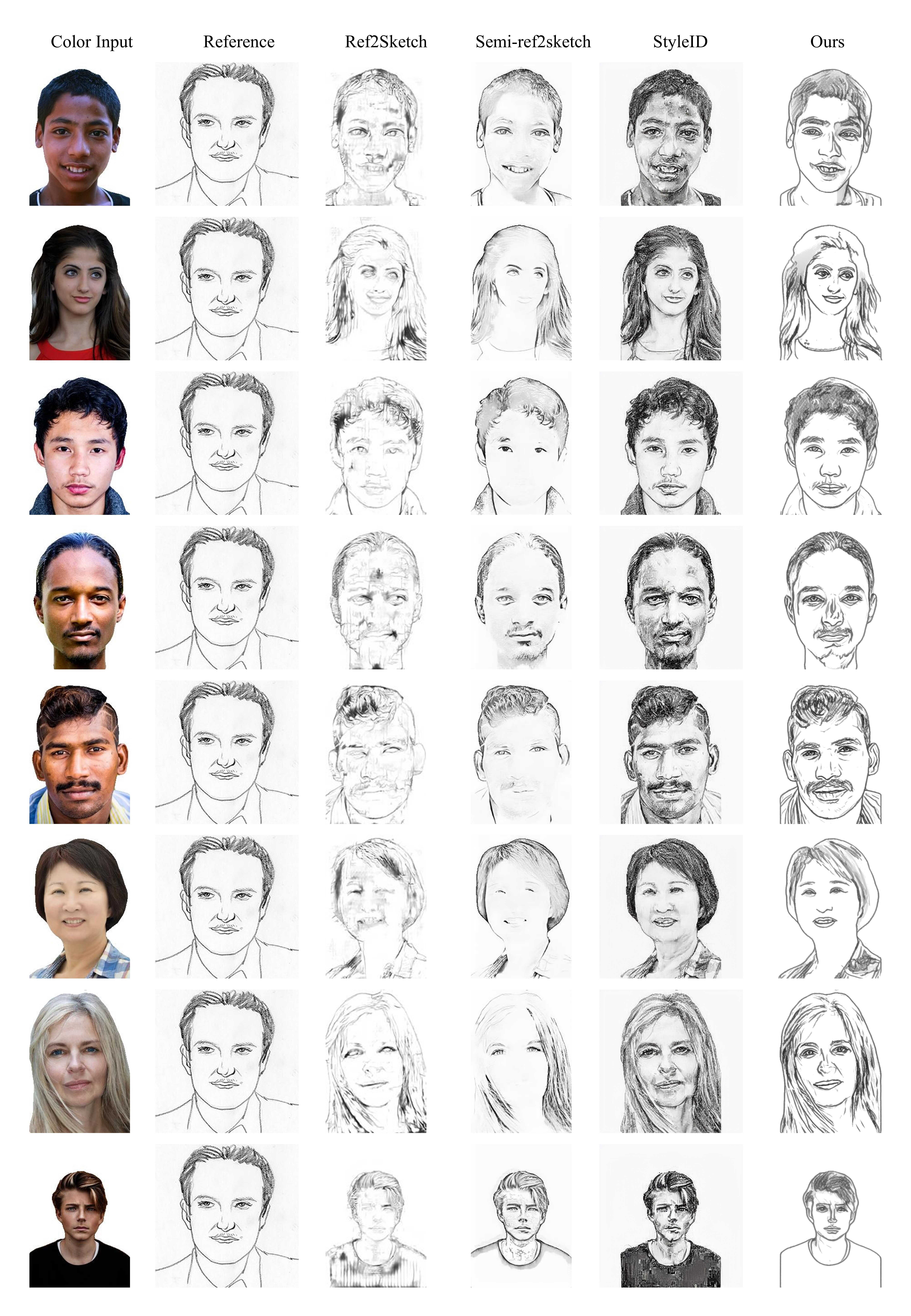}
  \caption{Final set of comparisons using the same reference sketch. Zoom in for viewing details.}
  \label{A8}
\end{figure*}

While our method outperforms both training-based and training-free approaches in quantitative evaluations, the results can vary depending on the reference sketch used. Although the 4SKST dataset clusters sketch images by style, testing different images within the same style can yield varying results. To address this, we have provided additional qualitative examples, comparing our method against Ref2Sketch, Semi-Ref2Sketch, and StyleID. These comparisons are illustrated in Figs. ~\ref{A3}--\ref{A8}.

We selected these methods due to their relevance: Ref2Sketch and Semi-Ref2Sketch are the top-performing training-based models for sketch generation, while StyleID represents a leading training-free style transfer method supporting reference-based sketch style transfer.

In qualitative comparisons, Semi-Ref2Sketch produced results comparable to ours on anime-style images, while Ref2Sketch performed slightly worse. Notably, both Ref2Sketch and Semi-Ref2Sketch are extensively trained on anime-style datasets, whereas our method operates without training. Despite this, our method not only performs well on anime-style images but also outperforms the others in user studies and in visual results for common sketch styles (e.g., brush or pen strokes). Even when quantitative differences (e.g., PSNR, FID, LPIPS) are minimal, our method shows superior performance. These findings reinforce the conclusion in the main paper that training-based sketch extraction methods struggle with unseen reference styles.

\begin{figure*}[t]
  \centering
  \includegraphics[width=0.9\linewidth]{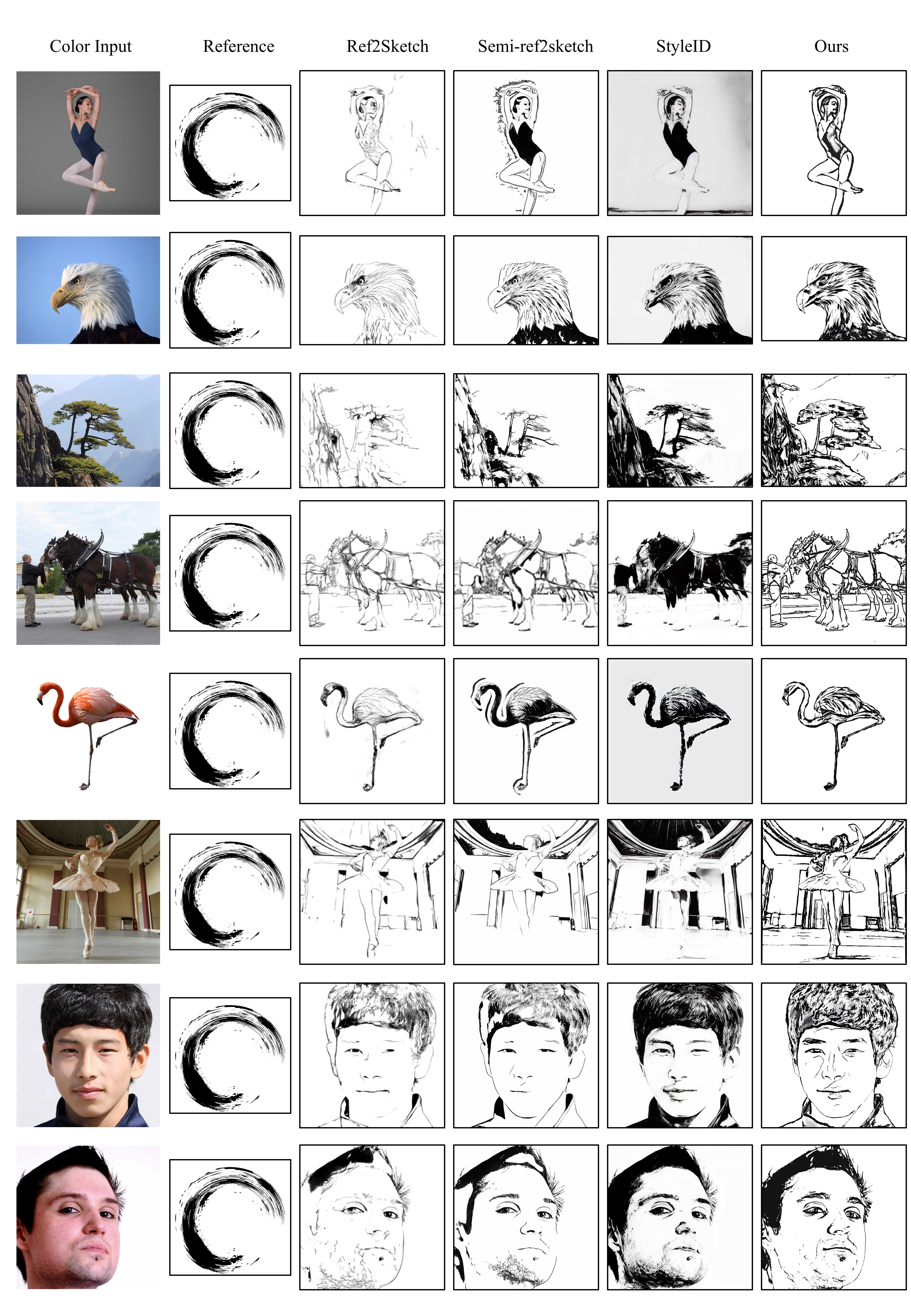}
  \caption{Comparison results for different models using a single reference sketch. Our model consistently outperforms state-of-the-art methods (Ref2Sketch, Semi-ref2sketch, and StyleID) in maintaining the reference sketch's brush strokes and artistic style across the input images. Zoom in to view details.}
  \label{A4}
\end{figure*}

\begin{figure*}[t]
  \centering
  \includegraphics[width=0.9\linewidth]{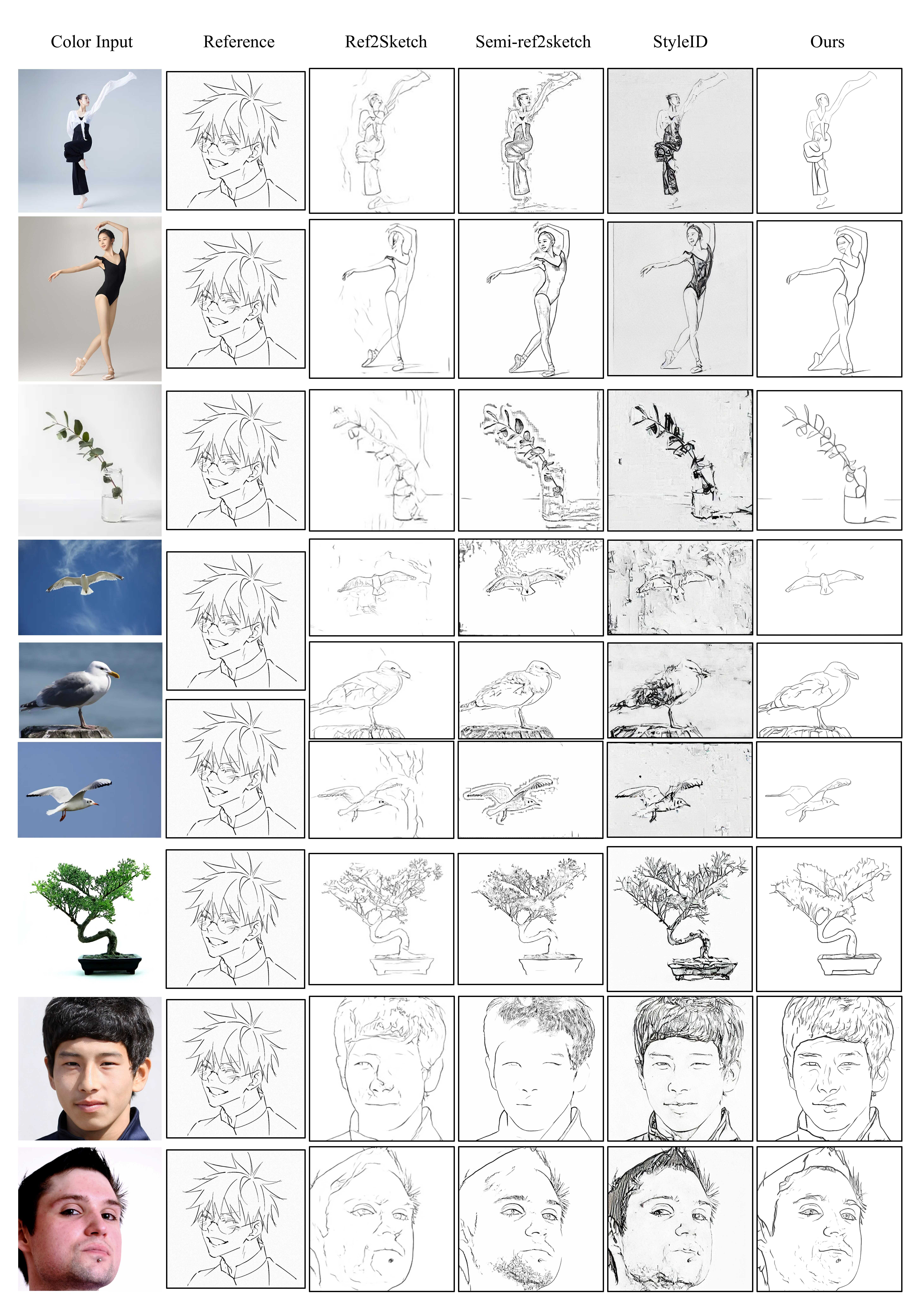}
  \caption{Continuation of comparison results with with baselines (Ref2Sketch, Semi-ref2sketch, and StyleID) using the same reference sketch. Zoom in for viewing details.}
  \label{A5}
\end{figure*}

